\PassOptionsToPackage{table, svgnames, dvipsnames}{xcolor}
\documentclass[fleqn,10pt]{wlscirep}
\usepackage{multirow}
\usepackage{anyfontsize}
\usepackage[caption = false]{subfig}
\usepackage{tabularx}
\usepackage{tabularray}
\usepackage[T1]{fontenc}
\usepackage{lastpage}
\title{Integrating Vision and Location with Transformers: A Multimodal Deep Learning Framework for Medical Wound Analysis
 }

\usepackage{makecell, cellspace, caption}

\author[1,*]{Ramin Mousa}
\author[2]{Hadis Taherinia}
\author[3,4]{Khabiba Abdiyeva}
\author[5]{Amir Ali Bengari}
\author[6]{Mohammadmahdi Vahediahmar}
\affil[1]{University of Zanjan, University Blvd., 45371-38791, Zanjan, I. R. Iran, Zanjan, 45371-38791, Iran.}
\affil[2]{Islamic Azad University Science and Research Branch,  Tehran, Iran}
\affil[3]{Samarkand State University named after Sharof Rashidov., University blv. 15., 140104., Samarkand., Uzbekistan}
\affil[4]{Kimyo International University In Tashkent., Shota Rustaveli street., 156., 100121., Tashkent., Uzbekistan}

\affil[5]{University of Tehran, 16th Azar St., Enghelab Sq., Tehran, Iran, Tehran, 1417466191, Iran.}
\affil[6]{Drexel University, Philadelphia, PA, USA}

\affil[*]{Raminmousa@znu.ac.ir}

\begin{abstract}
Effective recognition of acute and difficult-to-heal wounds is a necessary step in wound diagnosis. An efficient classification model can help wound specialists classify wound types with less financial and time costs and also help in deciding on the optimal treatment method. Traditional machine learning models suffer from feature selection and are usually cumbersome models for accurate recognition. Recently, deep learning (DL) has emerged as a powerful tool in wound diagnosis. Although DL seems promising for wound type recognition, there is still a large scope for improving the efficiency and accuracy of the model. In this study, a DL-based multimodal classifier was developed using wound images and their corresponding locations to classify them into multiple classes, including diabetic, pressure, surgical, and venous ulcers. A body map was also created to provide location data, which can help wound specialists label wound locations more effectively. The model uses a Vision Transformer to extract hierarchical features from input images, a Discrete Wavelet Transform (DWT) layer to capture low and high frequency components, and a Transformer to extract spatial features. The number of neurons and weight vector optimization were performed using three swarm-based optimization techniques (Monster Gorilla Toner (MGTO), Improved Gray Wolf Optimization (IGWO), and Fox Optimization Algorithm). The evaluation results show that weight vector optimization using optimization algorithms can increase diagnostic accuracy and make it a very effective approach for wound detection. In the classification using the original body map, the proposed model was able to achieve an accuracy of 0.8123 using image data and an accuracy of 0.8007 using a combination of image data and wound location. Also, the accuracy of the model in combination with the optimization models varied from 0.7801 to 0.8342.
\end{abstract}
\begin{document}

\flushbottom
\maketitle
%
%
\thispagestyle{empty}

\noindent Please note: Abbreviations should be introduced at the first mention in the main text – no abbreviations lists. Suggested structure of main text (not enforced) is provided below.

\section*{Introduction}

Chronic wound healing is a major issue in the world healthcare systems due to its uncountable findings that need an economic burden, complexity in presentation, and major impact on the patient's quality of life~\cite{huang2023development,patel2024integrated,anisuzzaman2022multi}. The valid and consistent classification of wounds correlates with proper diagnosis, care management, treatment regimens, and healed outcome forecasts~\cite{paper5}. Wound staging is predominantly still based on clinician experience, and its staging systems, namely the Meggitt-Wagner and University of Texas (UT) diabetic foot ulcer classification systems, are only moderate in interobserver reliability, indicating a need for integration with other clinical data~\cite{paper5}.

Conventional assessment of wounds is predominantly subjected to clinical examination; however, with the advent of computing sciences, there has been a tilt towards the use of computer vision and machine learning techniques. Several studies have proposed the deep learning approach for the classification of different wound types, including diabetic foot ulcers, pressure ulcers, venous ulcers, burns, and surgical wounds~\cite{anisuzzaman2022multi,huang2023development,paper11,patel2024integrated}. A few papers have also discussed differentiating various wound classes such as burns and pressure ulcers based on SVMs utilizing the pre-trained deep learning models with much encouraging results~\cite{paper14}.

In parallel to classification, the other advances in machine learning and artificial intelligence for wound evaluation concern non-invasive photo analysis techniques, including color correction, image segmentation, and texture feature extraction related to the wounds~\cite{wannous2008efficient,scebba2022detect}. Authors have recently reported that convolutional neural network models such as encoder-decoder networks and U-Net significantly boost automated analytical systems for wound segmentation~\cite{ wannous2008efficient}. Moreover, currently, mixed models that fuse wound image data with anatomical location information have shown superior performance compared to image-only approaches~\cite{anisuzzaman2022multi,patel2024integrated}. Standard anatomical body map tools have also helped create consistency in the reporting of locations and improved automated classifiers~\cite{patel2024integrated, anisuzzaman2022multi}.

Nonetheless, wound classification systems continue to exhibit variable performance due to differences within wound types, uncontrolled imaging conditions, and small datasets, and thus would benefit greatly from robust and generalizable methodology~\cite{patel2024integrated,scebba2022detect}. In addition, there is comparatively little AI implementation in certain types of wounds, especially infections at the surgical site of cardiac surgery.

The core focus of this study lies in the recent literature on automated wound classification and assessment. We propose an integrated deep learning architecture for wound classification with multimodal data comprising wound images and anatomical location information to achieve accurate and reliable wound classification in different clinical settings. The state-of-the-art techniques explored will potentially improve automated wound analysis and support clinical decision-making, thereby improving patient outcomes.

\section{Related Works}
One of the primary challenges in wound classification is the absence of a unified standard for diagnosing the type of wound and, consequently, selecting the most systematic and effective treatment method \cite{abazari2022systematic}. In this context, numerous studies have been carried out, and researchers have proposed various approaches for classifying and segmenting wounds based on images. Some of the proposed wound classification methods are based on physicians' and nurses' direct assessments. In these methods, experts are used to classify or label the data; for example, the evaluation of interobserver agreement in the Meggitt-Wagner and University of Texas (MW and UT) classification systems \cite{nagata2021skin} and the estimation of wound type based on human labeling in the STAR system \cite{santema2016comparing}. Although these methods are valuable due to the clinical knowledge of experts, they face some limitations compared to machine learning-based approaches, the most important of which are the limited volume of data and the dependence on the opinions of a small number of experts, which can affect the accuracy and generalizability of the final model.

Researchers have specifically studied diabetic ulcers as a type of wound using clinical data. \cite{huang2023development} developed a multi-task deep learning model based on convolutional neural networks (CNN) that could classify five types of wounds, including deep, infectious, arterial, venous, and pressure ulcers. The data used in this study were collected from Taipei Veterans General Hospital and included 2149 images from 1429 patients. The proposed architecture had task-specific attention branches to perform multiple classification objectives simultaneously. In the preprocessing stage, techniques such as manually cropping the images into squares, resizing to 300 × 300 pixels, data augmentation (by rotating, changing brightness and contrast), and using a weighted loss function to deal with class imbalance were applied. The model was evaluated using accuracy, sensitivity, specificity, area under the curve (AUC), and Cohen’s Kappa. The results indicated that the proposed model performed as well as, or in some cases better than, human experts and could be used by non-experts. However, the model relies solely on image data, and its generalizability needs to be further investigated in larger populations.
In a related study, \cite{huang2022image} developed a Fast R-CNN-based model with transfer learning (using Inception V2 and ResNet101) to classify diabetic foot ulcers, including vascular occlusion, suture wounds, and ulcerative wounds. The data were collected from Taichung Rongmin Hospital and expanded to 3600 images with data augmentation. Preprocessing included rotation, inversion, distortion, labeling, segmentation with GrabCut, and matching with SURF. The model achieved an accuracy of 89\% and an mAP of 87 in the best case, with a processing speed reported between 58 and 395 ms. In addition to real-time wound identification, this system also provides the ability to measure the area and perimeter of the wound. It can be used as a web-based diagnostic tool. However, the randomness of the GrabCut algorithm and the lack of coverage of some types of wounds are limitations of this method.

Another standard method for wound classification is classical machine learning algorithms such as SVM and Random Forest. In this regard, \cite{wannous2008efficient} classified pressure and diabetic ulcers based on tissue type (including granulation tissue, precursors, necrosis, and healthy skin) using an SVM algorithm and regional features. The data used was a private collection of images labeled by health professionals. Pre-processing steps included color correction using a Macbeth chart, unsupervised segmentation (JSEG), and color and texture feature extraction. The innovation of this research lies in combining color and texture features with regional segmentation methods and comparing its performance with other segmentation techniques. The proposed model achieved an accuracy of 94.5\% in classification and an average overlap score of 75.7\%, which showed a better performance than the pixel-based methods and even better than some experts' evaluations. Although the model's accuracy in detecting necrotic tissue was lower due to the limited training data and was dependent on high image quality, it was designed as part of the ESCALE project to develop a 3D wound assessment tool for clinical applications.

\cite{chitra2022investigation} presented a model based on a Random Forest (RF) algorithm to classify chronic wounds. The pre-processing involved unsupervised segmentation (J-SEG), mean shift filtering, texture extraction using GLCM, and color descriptors such as MCD, DCD, and histograms. The RF model surpassed both the neural network (88.08\%) and SVM (87.37\%) with an accuracy of 93.8\% and was designed as an automated diagnostic aid system for chronic wound tissue classification, showing potential for use in remote clinical settings. However, detailed information on the size and characteristics of the datasets is not provided, and comparisons with public databases or real environments are limited.
Another study developed a model based on the SVM algorithm and GLCM texture feature extraction using the combined data obtained from RSUD KRT Setjonegoro Hospital in Indonesia and the Kaggle database \cite{mawarni2023medical}. This study extracted features such as contrast, correlation, energy, and uniformity from the segmented wound areas and fed them into the SVM model to classify external wounds. Preprocessing was also performed by converting RGB images to grayscale, applying Active Contour segmentation, and normalizing the features. This model performed satisfactorily, achieving 96.39\% accuracy, 93.06\% classification accuracy, 92.85\% recall, and 92.58\% F1 score, and was proposed as an effective tool for early wound diagnosis in medical applications. However, limitations such as the small size of the data set and focusing only on the four predefined types of wounds have been reported for this method.

\cite{wannous2010enhanced} segmented wound tissues using color and texture features, multi-view imaging, and an SVM classifier. This method separated necrotic, granulation, and precursor tissues with relatively high accuracy. One advantage of this model is its ability to accurately identify different wound tissues. However, the limitation of the classification to three types of tissues and the dependence of the model on the quality of the three-dimensional image and the viewing angle are among its challenges.
Using the combination of RetinaNet models for detection and models such as U-Net, DeepLabV3, FCN, and ConvNet for segmentation, \cite{scebba2022detect} presented an approach called the detect-and-segment pipeline for chronic wound separation. This method was applied to public and private databases, including SW-DFU, SW-SSD, Medtec, SIH, and FUSC. The results showed that the proposed model achieved an MCC coefficient of 0.85 and an IoU index of up to 0.83 in the SW-DFU database. In many cases, its performance equaled or exceeded manual labeling. One of the most essential advantages of this method is its high generalizability to out-of-distribution data, maintaining performance even with less training data, and the possibility of using it with minimal user intervention. However, performance reduction in low data conditions (such as SW-SSD), the risk of incomplete or wrong detection, and limitations in the diversity of skin color and ethnicity in the training data are among the challenges of this model.

In recent years, thanks to the remarkable progress of artificial intelligence-based tools, the use of more advanced models, such as multi-modal models and explainable models in therapeutic applications, has become popular. Multi-modal models can provide more accurate classification of wounds by simultaneously utilizing different types of data, such as wound images and information such as their location. In contrast, explainable models provide users, especially clinicians, with a better understanding of the system's performance by explaining the model's decision-making process and reasons for choosing each class. \cite{anisuzzaman2022multi} introduced a multimodal model for wound classification that combines the wound image and its spatial information on a body map. The proposed architecture utilizes late-stage fusion of visual features extracted by convolutional neural networks alongside spatial data. The findings demonstrated that incorporating spatial information into the image significantly increased the classification accuracy compared to image-only models.

\cite{patel2024integrated} also presented a multimodal deep learning model that combined the wound image with its spatial information. In this model, a set of CNN networks, including VGG16, ResNet152, and EfficientNet-B2, for information fusion, along with axial attention modules and an adaptive gated MLP layer. Research showed that classification accuracy significantly improved when spatial data was incorporated, leading to enhanced model efficiency. \cite{Sarp2021} focused on increasing the interpretability of wound-classification models. This study used the transfer learning method using the VGG16 architecture, and XAI techniques, particularly LIME, were utilized to clarify the model's decisions. The model's output was presented in a graphical and comprehensible manner for clinicians, facilitating trust and usability for non-technical users. In another study, Lo et al. (2024) developed a deep learning-based model for vascular wound analysis that encompassed various tasks, including classification, depth estimation, and segmentation. Although the model relied solely on image data, it was designed with interpretability considerations in mind. For each prediction, confidence scores and explainability scores were provided, and modules were included to clarify the decision-making process within the model. Additionally, a web interface was developed to display the results in an understandable way for clinical users, which facilitates the application of this model in real environments.

A number of recent works have investigated deep learning-based automatic wound classification. Table \ref{tab:related_work} provides an overview of a few representative works with their data types, models, datasets, and major contributions.

In spite of the advancements, there exist various research gaps. Firstly, the majority of the studies consider only image data, whereas the fusion of multi-modal inputs, like the integration of images and wound location data, has been given very little consideration.

Second, although convolutional neural networks (CNNs) are prevalent, newer architectures such as Vision Transformers (ViTs) are relatively underexplored. Third, explainability of models is mostly restricted to simple tools such as LIME, with very few approaches providing holistic insight into the decision-making architectures. Lastly, a lot of the research is done on small or private datasets, which is dubious at best regarding the generalizability of the models to actual clinical practice. These limitations highlight the need to develop models that (i) utilize multi-modal inputs, (ii) leverage advanced architectures, (iii) enhance interpretability, and (iv) generalize well to a broad variety of datasets.

\begin{table}[htbp]
\centering
\resizebox{\textwidth}{!}{%
\begin{tabular}{|p{1.2cm}|p{2.8cm}|p{2.2cm}|p{2.2cm}|p{2.8cm}|p{2cm}|p{1.6cm}|p{2.6cm}|p{2.4cm}|p{2.4cm}|p{2.8cm}|p{2.6cm}|p{2.6cm}|}
\hline
\rowcolor{Gainsboro!60}
No. & Reference & Wound Type & Data Type & Model/Algorithm & Dataset & Dataset Size & Preprocessing & Evaluation Metrics & Best Performance & Innovation/Contribution & Advantages & Limitations \\
\hline
1 & Abazari et al., 2022 \cite{abazari2022systematic} & Chronic Wounds & Image & Deep CNN + Attention Mechanism & Private Dataset & 1077 images & Cropping, Augmentation & Accuracy, AUC & Accuracy 92.7\% & Attention-assisted feature extraction & Effective tissue classification & Limited dataset diversity \\
\hline
2 & Anisuzzaman et al. \cite{anisuzzaman2022multi}, 2022 & DFU & Image + Location & Late Fusion CNN & Private + Body Map & 1317 images & Normalization, Augmentation & AUC, Accuracy & AUC 0.85 & Multimodal fusion (Image + Location) & Boosts classification over unimodal & Requires annotated body maps \\
\hline
3 & Patel et al., 2024\cite{patel2024integrated} & Chronic Wounds & Image + Location & CNN (VGG16, ResNet152, EfficientNet) + Attention & Private Dataset & 1500 images & Augmentation, Scaling & Accuracy & Accuracy 84.8\% & Multimodal Transformer with Attention & Significant improvement & High model complexity \\
\hline
4 & Sarp et al., 2021\cite{Sarp2021} & Chronic Wounds & Image & VGG16 + LIME (XAI) & DFUC2020 (Public) & 2292 images & Resizing, Normalization & Accuracy, Explainability metrics & Accuracy 88.5\% & Integration of XAI for interpretability & Clinician-friendly explanations & Dataset imbalance \\
\hline
5 & Lo et al., 2024 & Vascular Wounds & Image & CNN + Depth Estimation + Explainability Module & AZH + Custom (Private) & 2957 images & Cropping, Augmentation & Accuracy, AUC, Explainability & Accuracy 90.1\% & Explainability Score Module & Real-time interface & Image-only limitation \\
\hline
\end{tabular}}
\caption{Summary of related works in wound analysis.}
\label{tab:related_work}
\end{table}

\section{Methodology}
\subsection{Dataset}
The AZH dataset was collected over two years from the AZH Wound and Vascular Center in Milwaukee, Wisconsin. It contains 730 de-identified wound images in JPEG format, each corresponding to one of four types of wounds: diabetic, pressure, venous, or surgical. The images were captured with an iPad Pro (iOS 13.4.1) and a Canon SX620 HS digital camera, and wound care specialists performed the labeling.
Most of the images in the dataset correspond to different patients. In cases where multiple images were taken from the same individual, they were taken from other locations in the body or at various stages of healing and treated as distinct samples due to visible differences in the shape and appearance of the wound. 

All images were anonymized by cropping to include only the wound and nearby skin, removing any identifiable information. The dataset is publicly available at \href {https://github.com/uwm-bigdata/Multi-modal-wound-classification-using-images-and-locations}{ GitHub}.

\begin{figure}[h]
  \centering
  \includegraphics[width=0.75\textwidth]{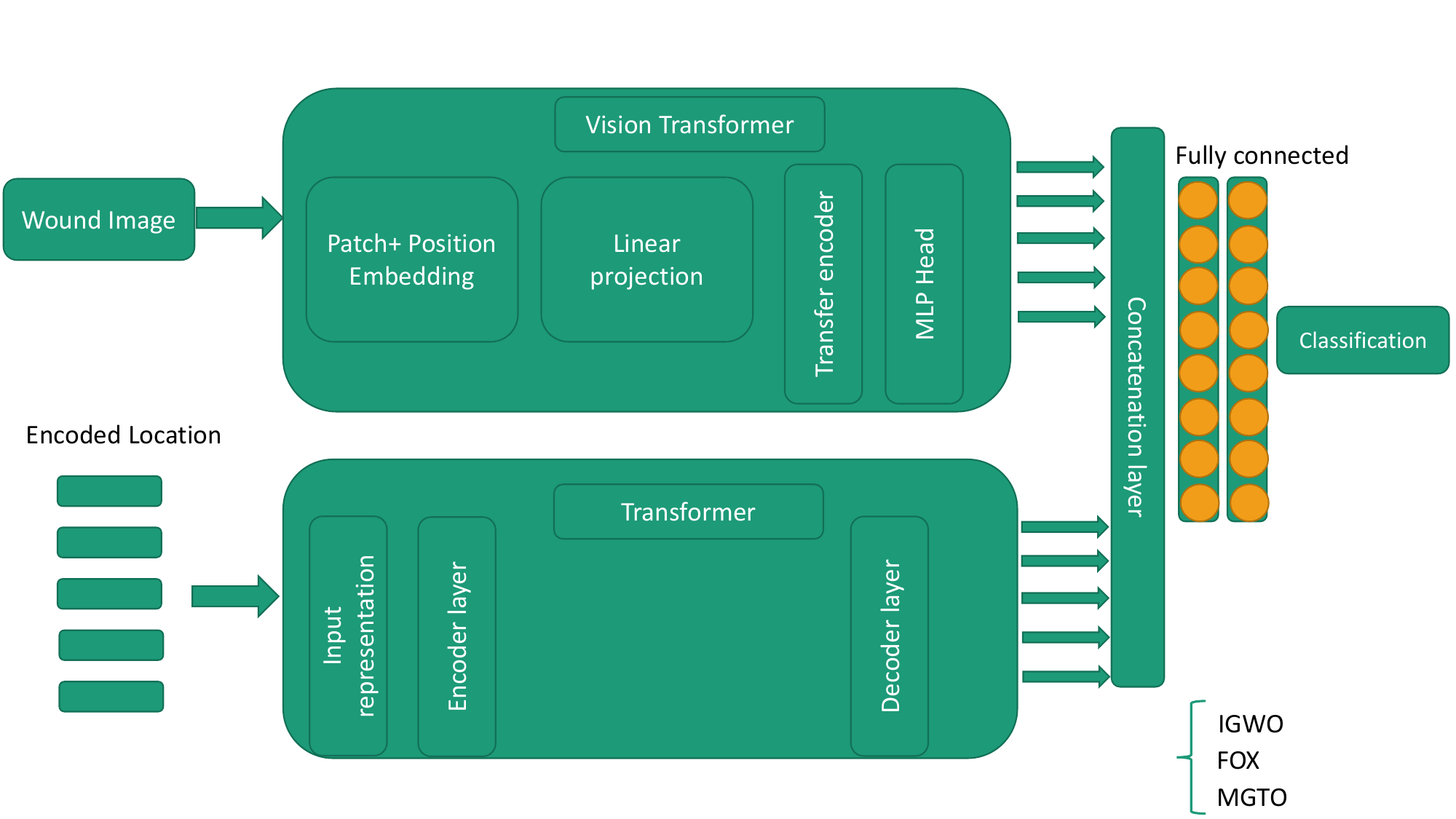}
  \caption{Wound multimodality classifer network architecture.}
  \label{fig:Wound}
\end{figure}

Figure \ref{fig:Wound} shows the architecture of the proposed wound-type classification network. This architecture uses two separate neural networks to work with image and location data. The final output is obtained by merging the outputs of the two networks. These two networks are merged in a concatenated layer. The following is a more detailed description of the components of these networks.

\subsection{Vision Transformers for Wound Image Classification}

Vision Transformers (ViTs) offer a tempting alternative to CNNs through global self-attention mechanisms instead of local receptive fields. This allows ViTs to represent complex spatial relationships in images of wounds for which multi-scale and context-sensitive features must be extracted.

\subsection*{Overview of Vision Transformer}

\begin{figure}[h]
  \centering
  \includegraphics[width=0.75\textwidth]{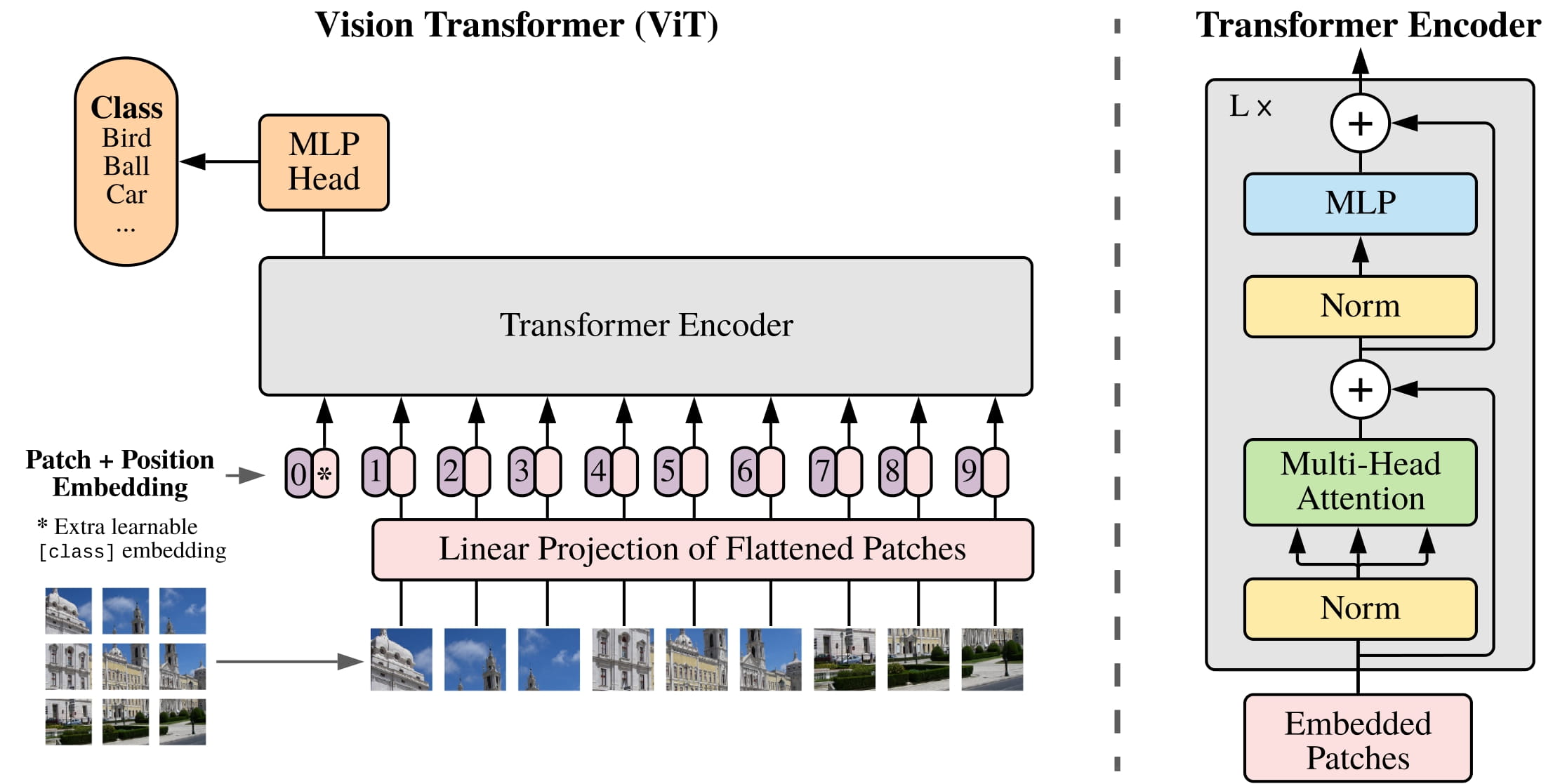}
  \caption{Model overview of the Vision Transformer (adapted from~\cite{dosovitskiy2020}).}
  \label{fig:vit_architecture}
\end{figure}

A comprehensive explanation of the Vision Transformer (ViT) architecture can be found in seminal works by Dosovitskiy et al. \cite{dosovitskiy2020} and Vaswani et al. \cite{vaswani2017attention}, with additional empirical evaluations provided in studies such as \cite{chen2021empirical}. Figure~\ref{fig:vit_architecture} displays the canonical pipeline of the Vision Transformer.

\textbf{Patch Embedding:} The input image is divided into a series of nonoverlapping patches. Each patch, denoted as \(x_p^i\) for the \(i^\text{th}\) patch (\(i \in \{1, \dots, N\}\)), is flattened and projected into a \(D\)-dimensional latent space via a trainable linear mapping \(\mathbf{E} \in \mathbb{R}^{P^2C \times D}\). To incorporate spatial information, these patch embeddings are augmented with a learnable positional embedding matrix \(\mathbf{E}_{\text{pos}} \in \mathbb{R}^{N \times D}\), resulting in:
\[
\mathbf{z}_0 = \left[ x_p^1 \mathbf{E};\, x_p^2 \mathbf{E};\, \dots;\, x_p^N \mathbf{E} \right] + \mathbf{E}_{\text{pos}}.
\]

\textbf{Transformer Encoder:} The sequence \(\mathbf{z}_0\) is processed through a stack of \(L\) identical transformer encoder layers. Each layer comprises a Multi-Head Self-Attention (MSA) module and a Multi-Layer Perceptron (MLP), interleaved with layer normalization and residual connections:
\[
\begin{aligned}
\mathbf{z}_\ell' &= \text{MSA}(\text{LN}(\mathbf{z}_{\ell-1})) + \mathbf{z}_{\ell-1}, \\
\mathbf{z}_\ell  &= \text{MLP}(\text{LN}(\mathbf{z}_\ell')) + \mathbf{z}_\ell',
\end{aligned}
\]
with the final class token being used for downstream classification.

\subsection*{Wavelet-Augmented Vision Transformer Architecture}

To enhance feature representation, particularly for medical textures like wounds, we propose an extension of the standard ViT with wavelet-based processing. The primary motivation is to inject multi-resolution spatial frequency information into the patch embeddings or attention heads. Specifically, we apply a discrete wavelet transform (DWT) to each image prior to patch embedding and integrate wavelet coefficients into the token representation. This retains both low- and high-frequency components that are crucial in wound edge detection, tissue granularity, and illumination variance.

The modified patch embedding becomes:
\begin{equation}
x_p^{i'} = \phi_{\text{DWT}}(x^i) \quad \text{then} \quad \mathbf{z}_0 = [x_p^{1'} \mathbf{E}; \ldots; x_p^{N'} \mathbf{E}] + \mathbf{E}_{\text{pos}},
\end{equation}
where $\phi_{\text{DWT}}$ denotes the wavelet transform operation. This hybrid architecture is especially suited for wound classification, where local texture and boundary patterns are diagnostically significant. ViTs capture global dependencies, while wavelet features enrich spatial precision, improving robustness to noise, blur, and variation in clinical imaging conditions. Recent works such as ~\cite{WooPark2024, Tanzi2021} demonstrate the effectiveness of Vits in surgical and trauma image analysis, further motivating our approach.
The output of applying Vit  on the input images is called $Vit_{latent}$. This vector contains the low-level features extracted from the image.
\subsection{Transformer}
\begin{figure}[h]
  \centering
  \includegraphics[width=0.75\textwidth]{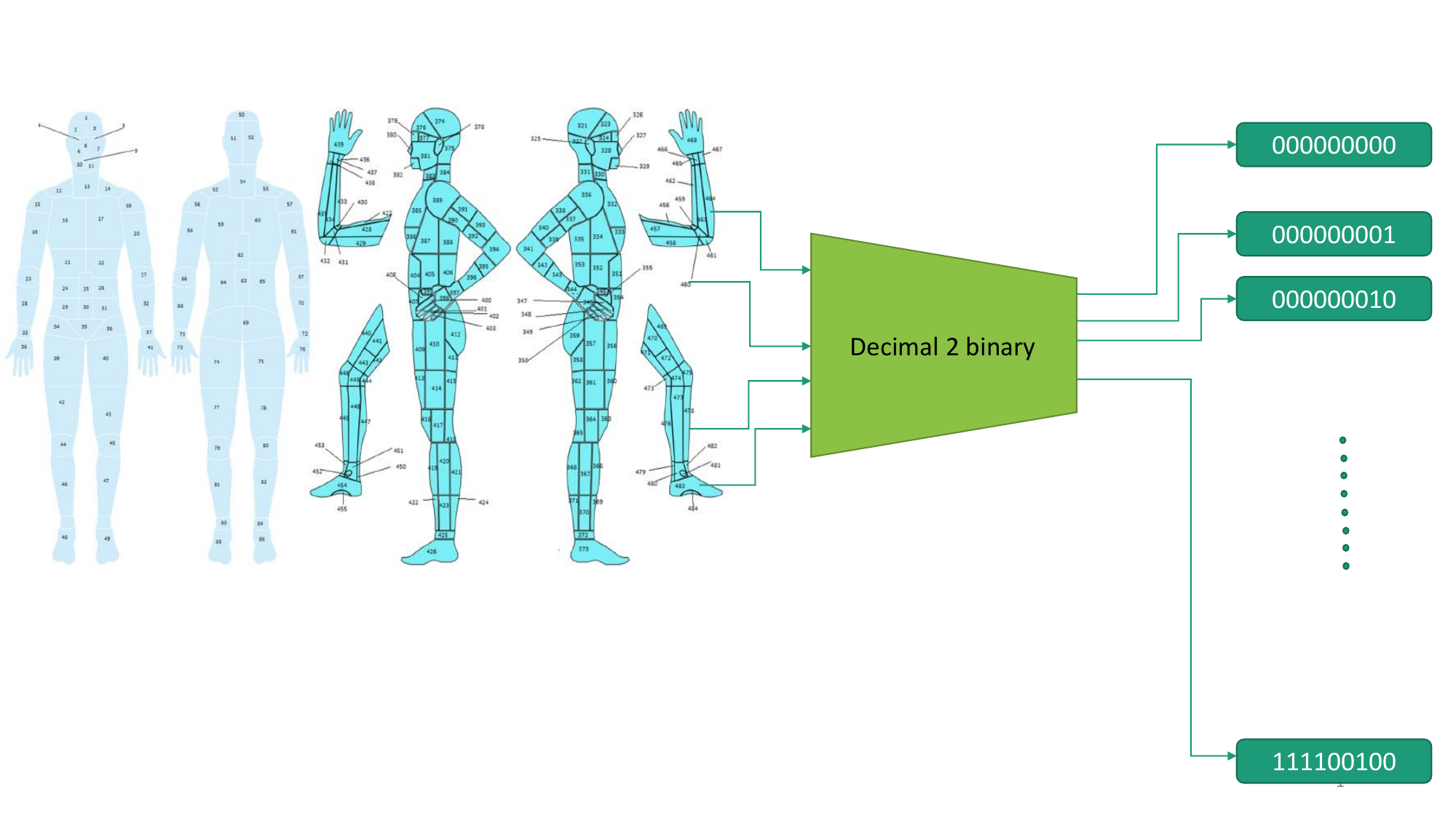}
  \caption{Decimal to binary converter process. The process of converting numbers related to the location of wounds from decimal to binary.}
  \label{fig:DEc}
\end{figure}

\begin{figure}[h]
  \centering
  \includegraphics[width=0.75\textwidth]{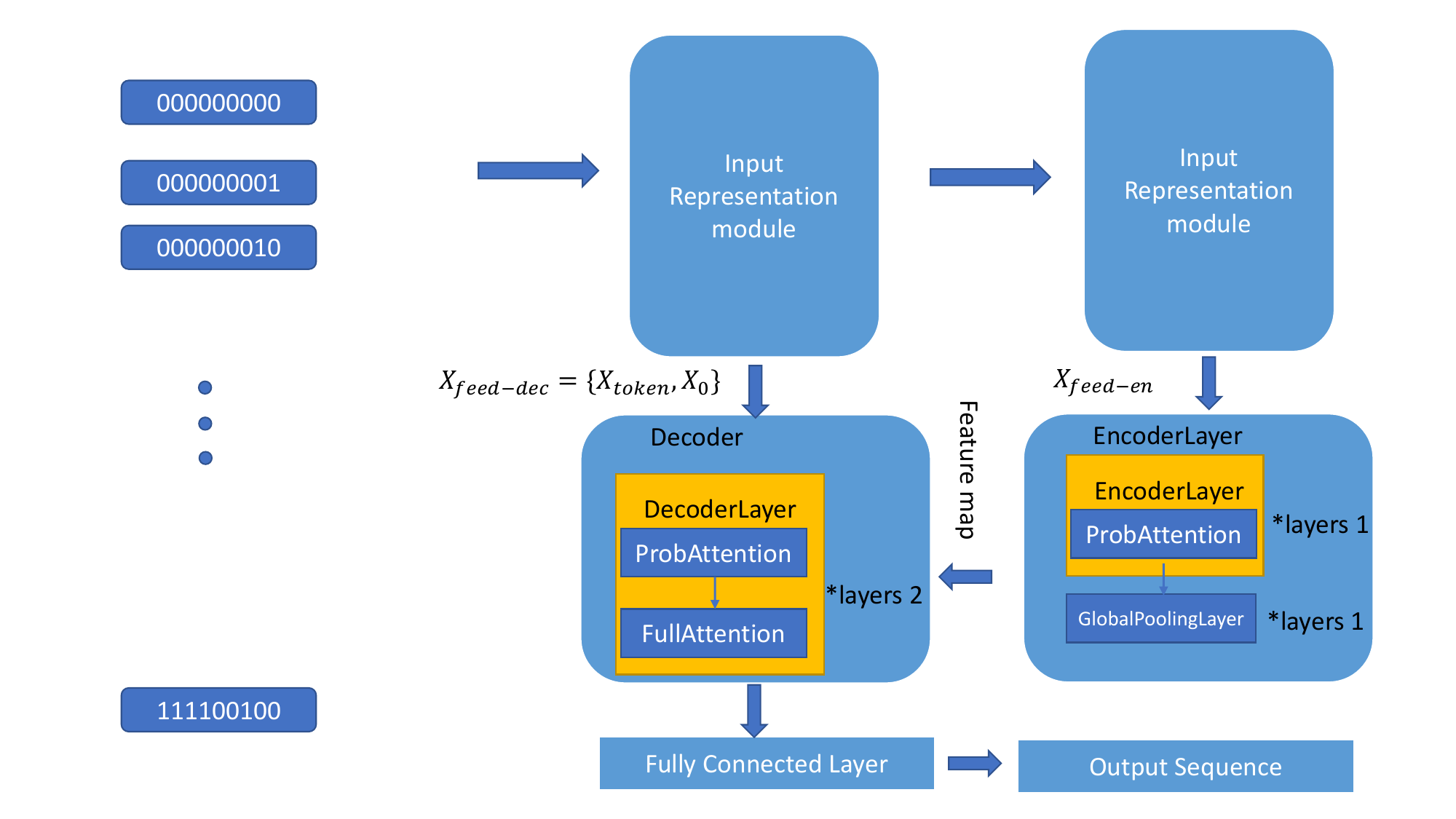}
  \caption{An overview of the transformer model for location encoded data.}
  \label{fig:Trans}
\end{figure}

The dataset consists of numerical features assigned to each body location. In the preprocessing process, as shown in Figure \ref{fig:DEc}, the raw numerical features were converted to binary numbers. This transformation, called DEC, converts each location feature into a 9-bit vector. These vectors are considered inputs to the transformer. The results of this conversion are given as input to the transformer network (see Figure \ref{fig:Trans}).

In the literature, Transformer was first used for machine translation. It is a sequence-by-sequence model designed with an encoder-decoder configuration. It takes a sequence of words from the source language as input and then produces a translation into the target language\cite{vaswani2017attention}. Transformer models jointly pay attention to and encode ordered information using self-attention and situational coding techniques. These techniques keep the sequential information intact for learning while eliminating the classical concept of recursion \cite{hao2019modeling}.

The transformer is a multi-purpose model and can be adapted to handle inputs and outputs related to different interpretations. The schematic diagram of the proposed transformer is shown in Figure \ref{fig:Trans}. Both the encoder and decoder accept an input. The encoder input is processed and combined with the decoder input using an attention mechanism. The encoder receives the appropriate input, while the decoder takes the past output values to track the past elements\cite{policarpitransformers}.
The main components of the transformer model are listed below:
\begin{itemize}
  \item Encoder Block: The encoder consists of a stack of N = 6 identical layers, each layer having two sub-layers. The first sub-layer is a multi-head self-attention mechanism, and the second sub-layer is a fully connected and simple feed-forward network. Also, around each of the two sub-layers, a residual connection should be used, and then the normalization of the layer should be done. That is, the output of each sublayer is $LayerNorm(x + Sublayer (x))$, where sublayer (x) is a function implemented by the sublayer itself. To facilitate these remaining connections, all model sublayers and embedded layers produce outputs with dimensions of dmodel = 512.
  \item Decoder Block: The decoder consists of a stack of N = 6 layers with the same structure. In addition to the two sub-layers in each encoder layer, the model introduces a third sub-layer to apply multi-head attention to the output of the encoder stack.
  \item Scaled Dot-Product Attention:
Here, the query along with the keys are divided by $\sqrt{d_k}$, then a Softmax operation is applied on them to determine the weight of the values. In practice, the attention function is computed on a set of queries simultaneously, packed together in a $Q$ matrix. Keys and values are also packed together in $K$ and $V$ matrices. 
Given an input sequence of binary wound locations represented as a vector $X$ with dimensions $[1:9]$, the attention mechanism computes a set of attention scores $A$ for each location $i$ in the sequence as follows:
\end{itemize}
\begin{enumerate}
    \item  Three new matrices are generated: 
    \begin{itemize}
            \item $Q_i$  Query matrix.
            \item  $K_i$ Key matrix.
            \item $V_i$ Value matrix,
        \end{itemize}

    \item Use dot product to compute attention scores between the query $Q_i$ and all key positions in the sequence.

    \[
    A_i = \text{softmax}\left(\frac{Q_i K^T}{\sqrt{d_k}}\right)
    \]

    \begin{itemize}
        \item $A_i$: Attention scores for position $i$.
        \item $Q_i$: The query matrix for position $i$.
        \item $K^T$: The transpose of the key matrix.
        \item $d_k$: The dimension of the key vectors,.
    \end{itemize}

    \item Using attention scores to calculate the weighted sum of value matrices:

    \[
    O_i = A_i V
    \]

    \begin{itemize}
        \item $O_i$: output vector.
        \item $A_i$: Attention scores.
        \item $V$: Value matrices.
    \end{itemize}

\end{enumerate}

This representation, where attention scores are assigned, places information at all locations, contributing to the current position's representation allowing the model to focus on the more relevant parts of the input while encoding information. The attention works by calculating scores based on comparing queries with keys to see the relative importance of various positions in the input string. This way, relationships and dependencies are captured, and the model can understand the contextuality and relationships of the data \cite{lin2022survey}. The mathematical formulation for the attention mechanism is written as follows:

\[
\text{Attention}(Q, K, V) = \text{softmax}\left(\frac{QK^T}{\sqrt{d_k}}\right)V
\]

Multi-Head Attention:
Instead of implementing a single attention function with dmodel dimensional keys, values, and queries, the authors found that linearizing the queries, keys, and values h times with different learned linear predictions of dk, dk, and dv dimensions, respectively, would be more beneficial. They executed the attention function in parallel on each of these predicted versions of the queries, keys, and values, and obtained the following dv output values:
\begin{equation}\label{b}
  MultiHead(Q,K,V)=Concat(head_1,\ldots,head_h ) W^O
\end{equation}
where
\begin{equation}\label{d}
head_i=Attention(QW_i^Q,KW_i^K,VW_i^v )
\end{equation}

The output of  transformer  is called $Transformer_{latent}$. 
At the end, the combination of $Vit_{latent}$ and $Transformer_{latent}$ are combined in a concatenate layer: 
\begin{equation}\label{d}
  Finall_{vector}=Vit_{latent} \bigoplus Transformer_{latent}
\end{equation}

\subsection{Optimization}
Three optimization approaches were used to optimize the network parameters. The details of these approaches are described below.

\subsubsection{Improved Grey Wolf Optimizer (IGWO)}

The Improved Grey Wolf Optimizer (IGWO) is a metaheuristic algorithm inspired by the group hunting behavior of grey wolves. In the basic version, the Grey Wolf Optimizer (GWO)\cite{kaveh2018improved}, wolves are categorized into four groups: Alpha, Beta, Delta, and Omega. The Alpha wolf, as the group leader, represents the best solution found so far, while Beta and Delta are advisors to the leader, and the other members (Omega) follow them to search for prey, or equivalently, the optimal solution. The hunting behavior is modeled through three phases: searching for prey, encircling prey, and attacking\cite{kaveh2018improved}.

The encircling behavior is mathematically modeled as\cite{li2021improved}:

\begin{equation}
\vec{D} = \left| \vec{C} \cdot \vec{X}_p(t) - \vec{X}(t) \right|
\end{equation}
\begin{equation}
\vec{X}(t+1) = \vec{X}_p(t) - \vec{A} \cdot \vec{D}
\end{equation}

where \( \vec{A} \) and \( \vec{C} \) are coefficient vectors calculated as\cite{li2021improved}:

\begin{equation}
\vec{A} = 2a \cdot \vec{r}_1 - a, \quad \vec{C} = 2 \cdot \vec{r}_2
\end{equation}

and

\begin{equation}
a = 2 - 2 \frac{t}{\text{MaxIter}}
\end{equation}

To update positions relative to the three best solutions\cite{li2021improved}:

\begin{equation}
\vec{X}(t+1) = \frac{\vec{X}_1 + \vec{X}_2 + \vec{X}_3}{3}
\end{equation}

Improvements include the use of a Tent chaotic map:

\begin{equation}
y_{i+1} = (2y_i) \mod 1
\end{equation}

Gaussian Mutation and a Cosine Control Factor \cite{li2021improved}:

\begin{equation}
a' = 2 \cdot \cos\left( \frac{\pi}{2} \cdot \frac{t}{\text{MaxIter}} \right)
\end{equation}

\subsubsection{Fox Optimization Algorithm (FOX)}

The Fox Optimization Algorithm (FOX) is a metaheuristic algorithm inspired by the natural hunting behavior of foxes. In this method, the fox's movement is modeled based on measuring the distance to the prey and performing efficient leaps toward it. This algorithm possesses strong local search capabilities and effectively avoids entrapment in local optima. The FOX algorithm has been successfully applied to engineering design problems, including the design of pressure vessels. The sensitivity analysis of the algorithm's parameters shows that FOX can maintain a good balance between exploration and exploitation during optimization\cite{mohammed2023fox}.

At the beginning, FOX initializes the population \(X\), where each individual represents a red fox's position. The fitness of each individual is evaluated, and the best solution is selected\cite{mohammed2023fox}.

\subsubsection*{Exploitation Phase}

\begin{equation}
\text{Dist}_{S\_T_{it}} = Sp_S \times \text{Time}_{S\_T_{it}}
\end{equation}

\begin{equation}
\text{Dist}_{Fox\_Prey_{it}} = \frac{1}{2} \times \text{Dist}_{S\_T_{it}}
\end{equation}

\begin{equation}
\text{Jump}_{it} = 0.5 \times 9.81 \times t^2
\end{equation}

Update position:

\begin{equation}
X_{it+1} = \text{Dist}_{Fox\_Prey_{it}} \times \text{Jump}_{it} \times c
\end{equation}

\subsubsection*{Exploration Phase}

\begin{equation}
tt = \frac{\sum \text{Time}_{S\_T_{it}}(i,:)}{\text{dimension}}
\end{equation}

\begin{equation}
\text{MinT} = \min(tt)
\end{equation}

\begin{equation}
X_{it+1} = BestX_{it} \times \text{rand}(1, \text{dimension}) \times \text{MinT} \times a
\end{equation}

\subsubsection{Improved Gorilla Troops Optimizer (mGTO)}

The Gorilla Troops Optimizer (GTO) is a metaheuristic algorithm developed based on gorillas' social behavior and foraging. In the initial version, the search process consisted of two main phases, i.e., exploratory search and local exploitation, which were modeled through the simulation of the social interactions of gorillas. However, the original GTO version faced challenges such as early convergence and entrapment in local optima. To overcome these limitations, an improved version, called mGTO, was presented\cite{mostafa2023improved} .

\begin{equation}
X(t+1) =
\begin{cases}
(Ul - Ll) \times r_1 + Ll & \text{if } \text{rand} < p, \\
(r_2 - C) \times X_r(t) + L \times H & \text{if } \text{rand} \geq 0.5, \\
X(t) - L \times (L \times (X(t) - X_r(t)) + r_3 \times (X(t) - X_r(t))) & \text{otherwise}.
\end{cases}
\end{equation}

Improvements include EOBL:

\begin{equation}
\tilde{x}_k = y_k + z_k - x_k
\end{equation}

CICD inverse formula:

\begin{equation}
F^{-1}(p; 0,1) = \tan\left( \pi (p - 0.5) \right)
\end{equation}

and TFO movement:

\begin{equation}
TFO = \tan\left( \frac{v \pi}{2} \right)
\end{equation}

\section{Experimental result}

The hardware and software specifications of the proposed model are given in Table \ref{TableSoft}.  To evaluate the proposed model, the following six evaluation criteria were used in classification and segmentation tasks:
\begin{enumerate}
    \item \textbf{Sensitivity(\(Se\)):} Proportion of true positives correctly identified: \(Se = \frac{a}{a+c}\)
    \item \textbf{Specificity (\(Sp\)): }Proportion of true negatives correctly identified: \(Sp = \frac{d}{b+d}\) 
    \item \textbf{Accuracy: }Fraction of correct predictions \(\text{Accuracy} = \frac{\text{TP} + \text{TN}}{\text{Total Observations}}\) 
    \item \textbf{Precision:} Proportion of true positives among predicted positives: \(\text{Precision} = \frac{\text{TP}}{\text{TP} + \text{FP}}\) 
    \item \textbf{Recall: }Proportion of true positives correctly identified: \(\text{Recall} = \frac{\text{TP}}{\text{TP} + \text{FN}}\)       
    \item \textbf{F1-Score:} Harmonic mean of precision and recall: \(\text{F1} = 2 \cdot \frac{\text{Precision} \cdot \text{Recall}}{\text{Precision} + \text{Recall}}\) 
\end{enumerate}
\begin{table*}[]
\centering
\caption{The hardware and software specifications of this research.}
\begin{tabular}{|lll|}
\hline
\rowcolor{Gainsboro!60}
\multicolumn{3}{|c|}{Software}                                                                              \\ \hline
\multicolumn{1}{|l|}{Name}                       & \multicolumn{1}{l|}{version} & Description               \\ \hline
\multicolumn{1}{|l|}{Ubuntu Bionic Beaver (LTS)} & \multicolumn{1}{l|}{18.04.2} & Operating System          \\ \hline
\multicolumn{1}{|l|}{Python}                     & \multicolumn{1}{l|}{3.12.3}   & Used for implementation   \\ \hline
\multicolumn{1}{|l|}{Keras}                      & \multicolumn{1}{l|}{2.15.0}   & Used for building models  \\ \hline
\multicolumn{1}{|l|}{Pandas}                     & \multicolumn{1}{l|}{0.23.4}  & Used for data analysis    \\ \hline
\multicolumn{1}{|l|}{Tensorflow}                 & \multicolumn{1}{l|}{v2.16.1}  & Used as backend for Keras \\ \hline
\multicolumn{1}{|l|}{CUDA}                       & \multicolumn{1}{l|}{9.0.176} & Required for Tensorflow   \\ \hline
\multicolumn{1}{|l|}{cuDNN}                      & \multicolumn{1}{l|}{7.4.1}   & Required for Tensorflow   \\ \hline
\rowcolor{Gainsboro!60}
\multicolumn{3}{|c|}{Hardware}                                                                              \\ \hline
\multicolumn{1}{|l|}{Name}                       & \multicolumn{2}{l|}{Version}                             \\ \hline
\multicolumn{1}{|l|}{CPU}                        & \multicolumn{2}{l|}{Intel i7-2600}                       \\ \hline
\multicolumn{1}{|l|}{GPU NVIDIA}                 & \multicolumn{2}{l|}{GeForce GTX 980}                     \\ \hline
\multicolumn{1}{|l|}{Memory}                     & \multicolumn{2}{l|}{Kingston 8 GB DDR3}                  \\ \hline
\multicolumn{1}{|l|}{GPU Memory}                 & \multicolumn{2}{l|}{16 GB, GDDR5}                         \\ \hline
\end{tabular}
\label{TableSoft}
\end{table*}

Four wound class classifications (D vs. P vs. S vs. V) on the AZH dataset were selected to select the best combinations for the proposed network. This classification was the most challenging combination, as there were no normal skin (N) or background (BG) images in the experiment. This experiment was conducted with the initially developed body map containing 484 locations. Table \ref{Table1} shows the results of this experiment. We also present the results on the original dataset (without any augmentation) for this experiment to show the impact (improvement) of the data augmentation.
In Table \ref{Table1}, three input modes, Location, Image, and Location+ Image, are considered as Original and Augmented data. In Location and Original data mode, the maximum accuracy of 0.7712 was obtained by the Transformer+ Binary encoding approach. The lowest value reported in these inputs is by MLP. In Augmented mode, the Transformer+ Binary encoding approach also obtained the best result. In Image input, the transfer models obtained different results depending on their structure. In the models reported in this, the highest accuracy obtained in both Original and Augmented modes was 0.6576 and 0.7173, which the VGG16 model recorded. In combining the transfer models with the Capsule approach, the best result was obtained by the EfficientNetB4 model in both modes. The Vit model in the original mode was able to achieve Accuracy=0.799, Precision=0.801, Recall=0.8045, and F1=0.8027. This model, combined with Augmented data, achieved Accuracy=0.8283. The Vit+ Wavelet model achieved maximum accuracy in both Original and Augmented modes. This model achieved an accuracy of 0.8123 in the Original mode and 0.8355 in the Augmented mode. In the Image+Location mode in the Original data, the tested models, combined with MLP, achieved a maximum accuracy of 0.7717. In combination with LSTM, these models were able to achieve a maximum accuracy of 0.7283. In this data, the Xception + GMRNN model achieved Accuracy=0.7877, Precision=0.7882, Recall=0.7797, and F1=0.7797. This model also achieved Specificity=0.9662 and Sensitivity=0.9334. The proposed Vit+Transformer model achieved very close results to Xception + GMRNN, achieving an accuracy of 0.7876 in the test data. 

The Vit+ Wavelet+ Transformer model obtained the best result from the original data. This model was able to achieve an accuracy of 0.8007. On the Image+ location dataset and in the Augmented data mode, the combination of models with MLP achieved a maximum accuracy of 0.78. In combination with LSTM, a maximum accuracy of 0.7935 was obtained. The Xception+ GMRNN model achieved an accuracy of 0.8189, and the Vit+ Transformer model also achieved an accuracy of 0.8188. These two models performed similarly. The Vit+ Wavelet+ Transformer approach obtained the maximum accuracy on this dataset. This model achieved an accuracy of 0.8354, an improvement of 0.0347 over the Original data.

\begin{figure*}
\centering
\subfloat[ Location]{\includegraphics[width=1\textwidth]{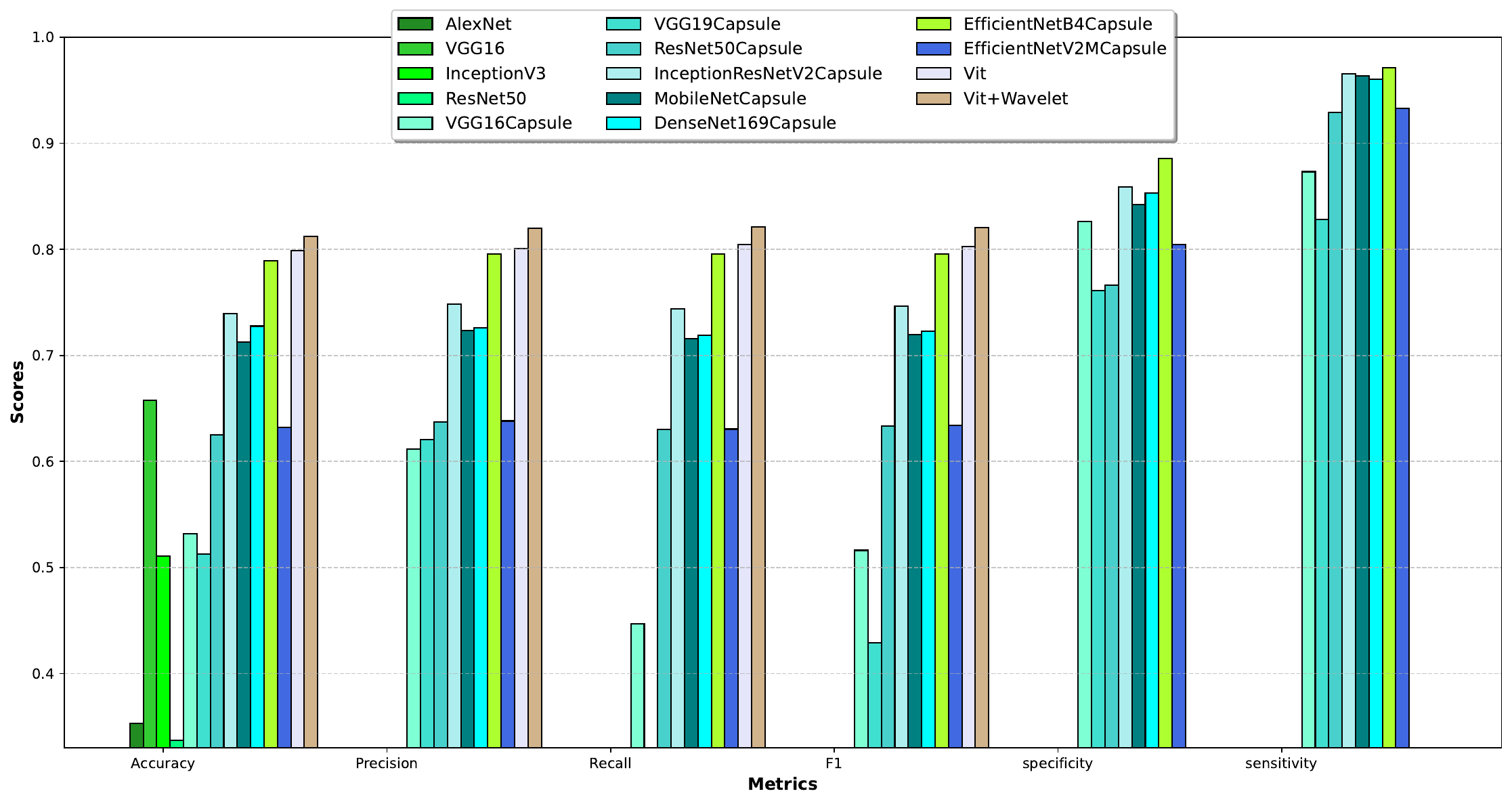}}\\
\subfloat[Image]{\includegraphics[width=1\textwidth]{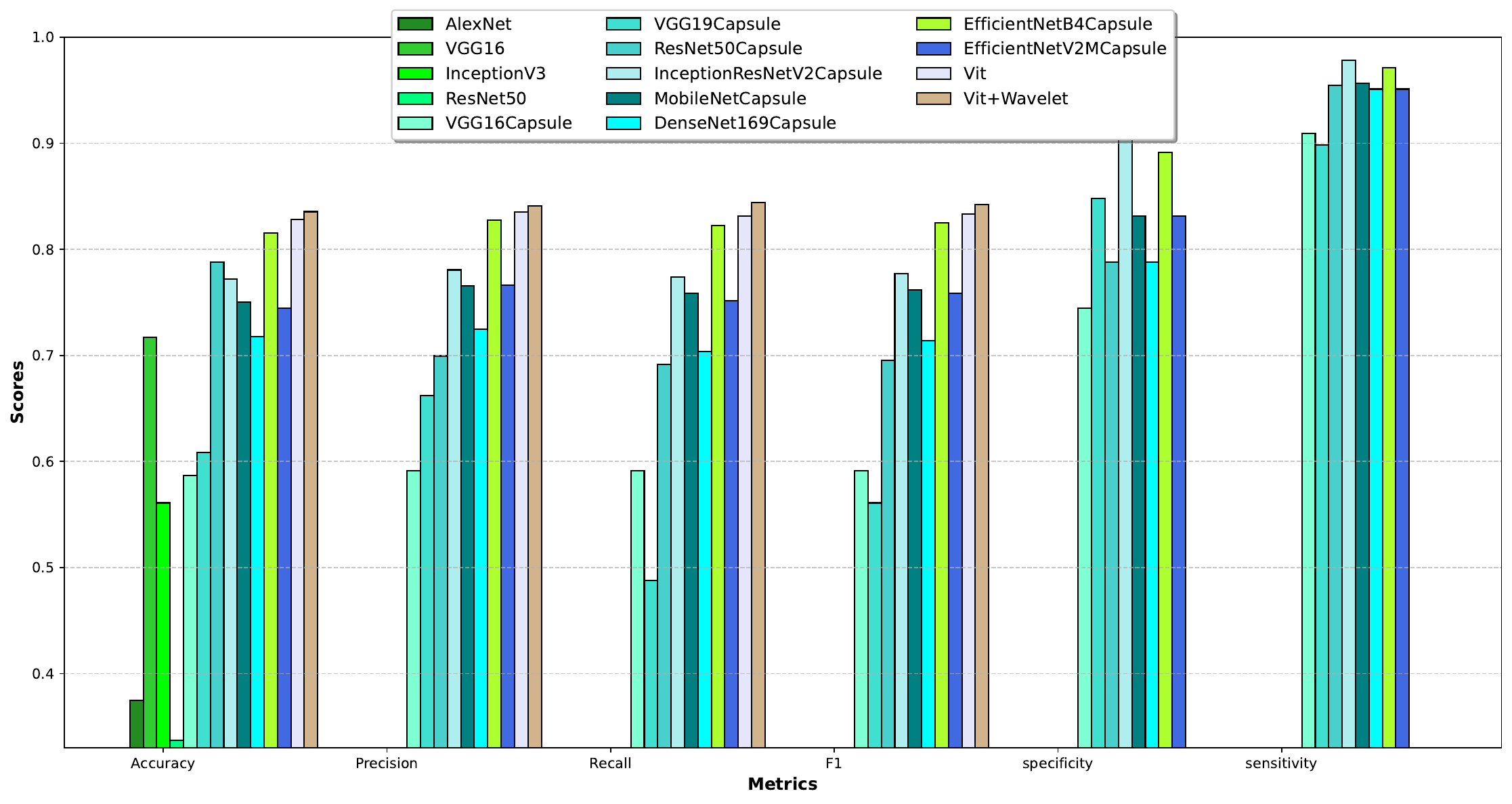}}\\
\caption{Bar plot for four wound class classification (D vs. P vs. S vs. V) on AZH dataset(Original Data).}
\label{fig:fig1}
\end{figure*}

Figure \ref{fig:fig1} illustrates the performance comparison of models on four-class wound classification (D vs. P vs. S vs. V) using the original AZH dataset with a simplified body map of 323 locations. In subplot (a), which uses Location input, ViT+Wavelet achieved the highest accuracy, precision, recall, and F1-score among all tested models. Capsule-based models like DenseNet169Capsule and EfficientNetB4Capsule also performed well, especially in terms of specificity and sensitivity. Models such as AlexNet and InceptionV3 recorded the lowest performance in this setting. In subplot (b), using Image input, the performance trend remained consistent.ViT+Wavelet consistently topped all important measures, with Capsule-boosted EfficientNet and DenseNet models ranking close second. AlexNet consistently registered the worst performance. These graphical representations validate the exemplary performance of ViT+Wavelet for both input modalities with respect to the original data set, with models with Capsule-based architecture also demonstrating competitive performance.

\begin{figure*}
\centering
\subfloat[ Original]{\includegraphics[width=1\textwidth]{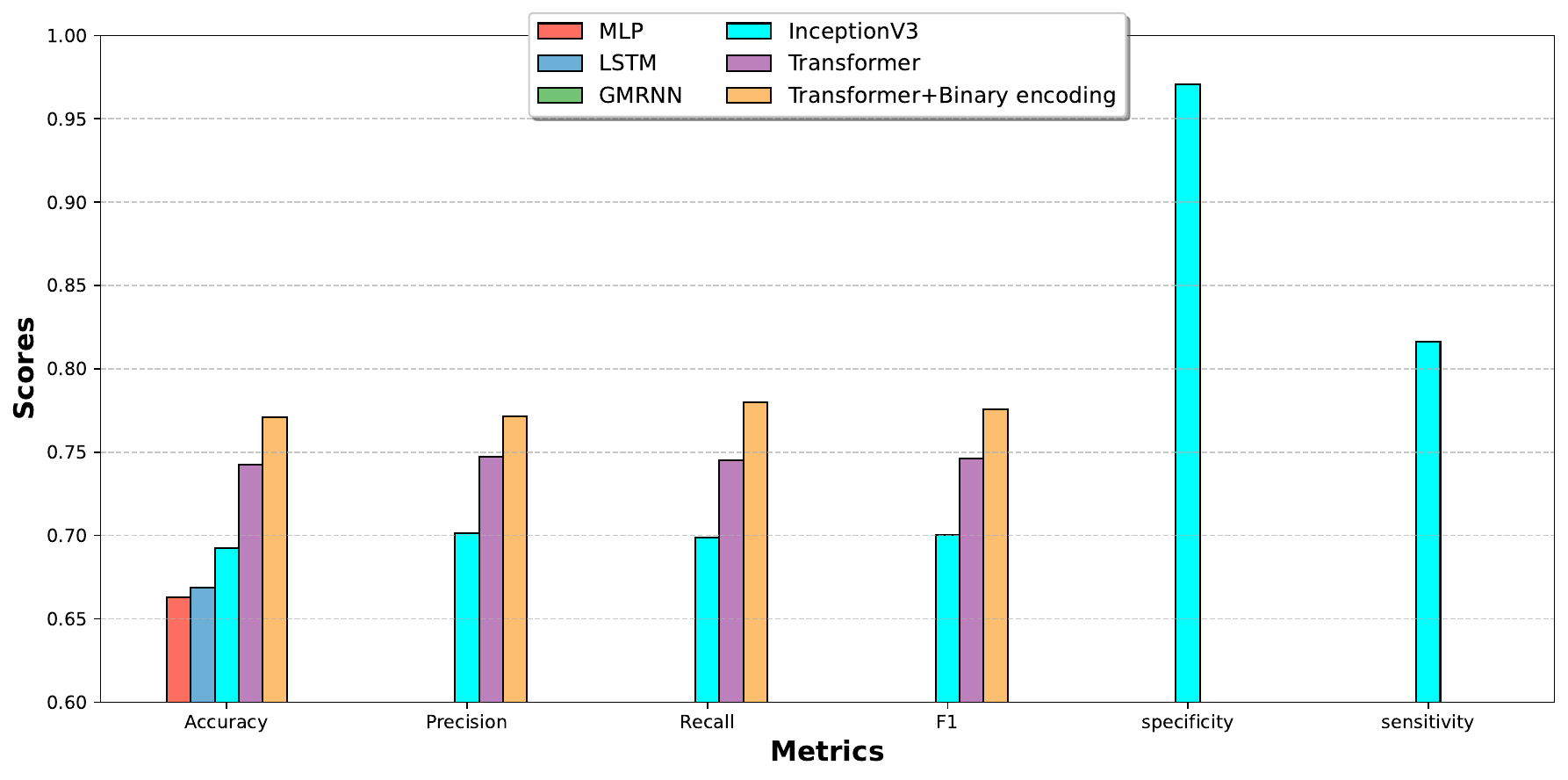}}\\
\subfloat[ Augmented]{\includegraphics[width=1\textwidth]{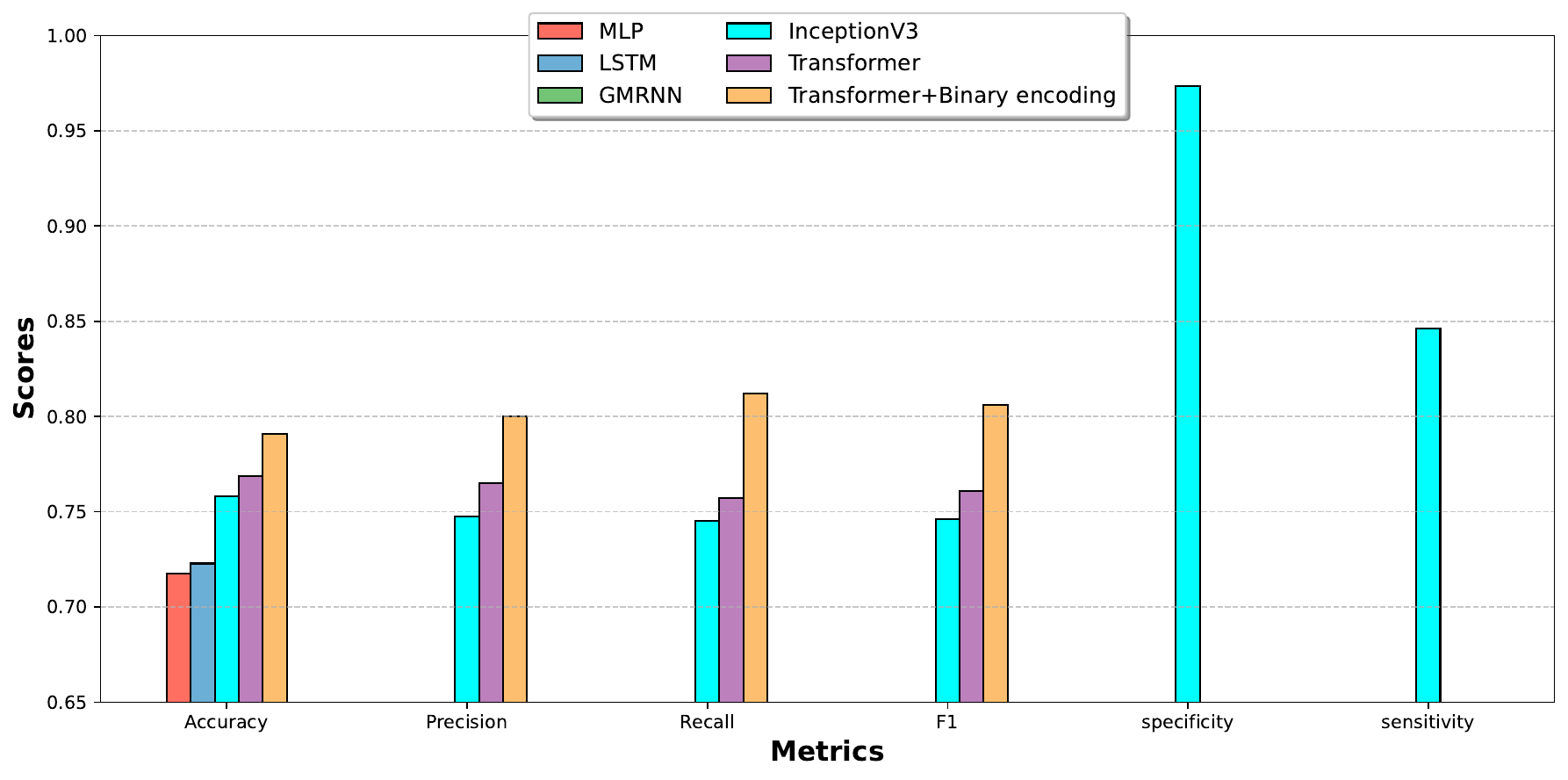}}\\
\caption{Bar plot for four wound class classification (D vs. P vs. S vs. V) on AZH dataset(Original and Augmented Data).}
\label{fig:fig3}
\end{figure*}

Figure \ref{fig:fig3} shows a performance comparison among different models using \textit{Image} inputs on the original and augmented datasets for the four-class wound classification problem (D vs. P vs. S vs. V). In both subfigures (a) and (b), the Transformer+Binary encoding method shows the best performance in most of the metrics, i.e., Accuracy, Precision, Recall, and F1-score. Within the Original dataset setup (Figure 3a), although InceptionV3 achieved a high degree of specificity, its sensitivity and overall performance measures were very low. By contrast, in the augmented setup (Figure 3b), Transformer-based models had significantly better performance than simpler models such as MLP and LSTM. Across both kinds of input, MLP consistently showed the worst-performing results on every metric considered. The findings highlight the superiority of Transformer-based models, especially when combined with binary encoding.

\begin{figure*}
\centering
\subfloat[ Location]{\includegraphics[width=1\textwidth]{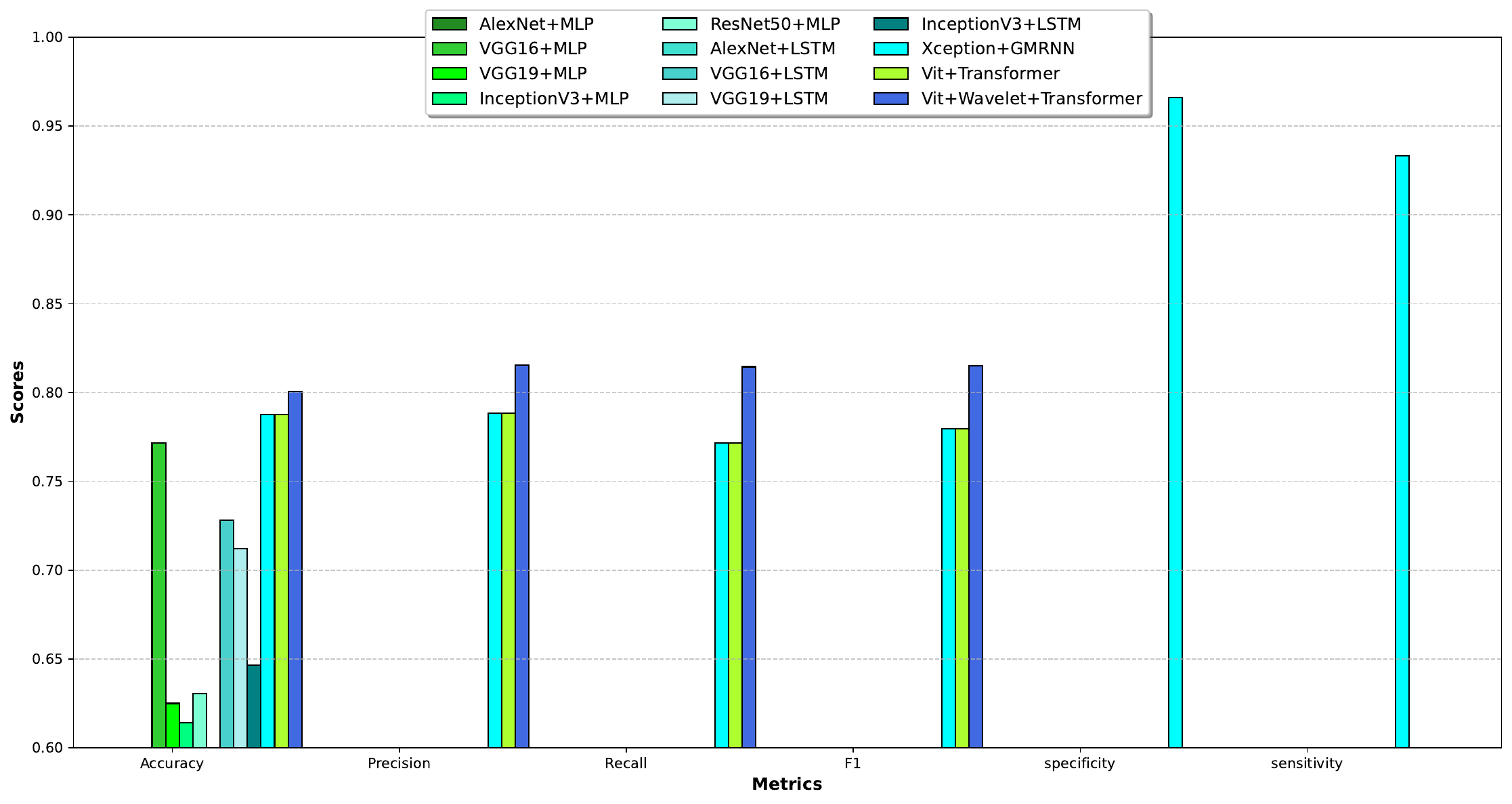}}\\
\subfloat[Image]{\includegraphics[width=1\textwidth]{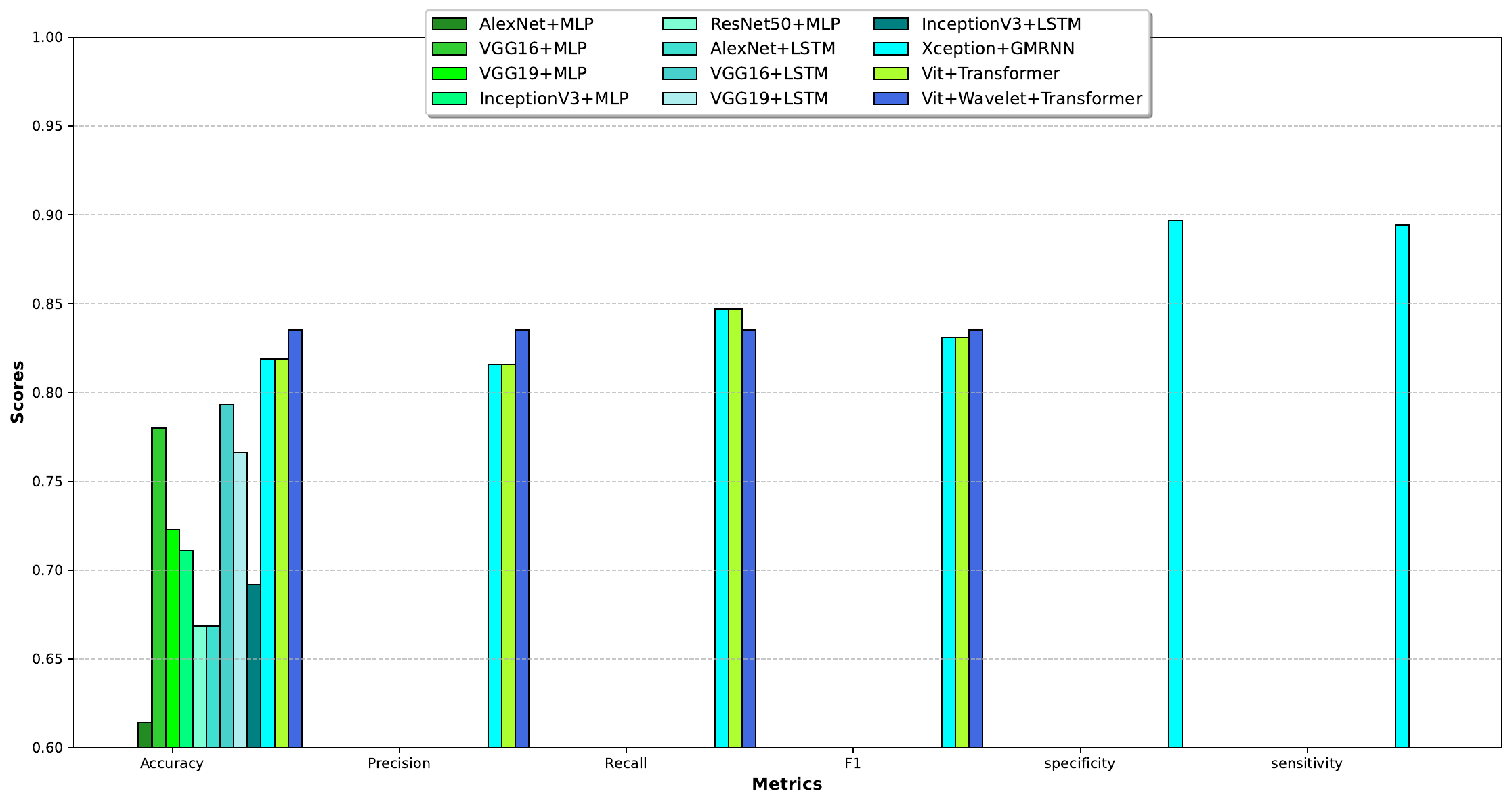}}\\
\caption{Bar plot for four wound class classification (D vs. P vs. S vs. V) on AZH dataset(Original Data).}
\label{fig:fig4}
\end{figure*}

Figure \ref{fig:fig4} presents the performance comparison of models for four-class wound classification (D vs. P vs. S vs. V) on the AZH dataset using augmented data. In subplot (a), using Location input, the ViT+Wavelet+Transformer model outperformed all others, achieving the highest accuracy, precision, recall, and F1-score. The Xception+GMRNN model also showed strong results, particularly in specificity and sensitivity. In subplot (b), with Image input, the ViT+Wavelet+Transformer again led in all core metrics, demonstrating its robustness across modalities. Compared to original data, most models showed improved accuracy after augmentation, confirming the effectiveness of data augmentation in enhancing model performance.

\begin{table}
\centering
\resizebox{\textwidth}{!}{
\begin{tabular}{|l|l|c|c|c|c|c|c!{\vrule width 1.2pt}c|c|c|c|c|c|} 
\hline
\rowcolor{Gainsboro!60}
~                & ~                                                                    & Accuracy                                             & Precision                                           & Recall                                              & F1                                                  & specificity                                         & sensitivity                                         & Accuracy                                             & Precision                                             & Recall                                               & F1                                                   & specificity                                          & sensitivity                                          \\ 
\hline
Location         & ~                                                                    & \multicolumn{1}{l}{}                                 & \multicolumn{1}{l}{}                                & \multicolumn{1}{l}{}                                & \multicolumn{1}{l}{Original Data}                   & \multicolumn{1}{l}{}                                &                                                     & \multicolumn{1}{l}{}                                 & \multicolumn{1}{l}{}                                  & \multicolumn{1}{l}{}                                 & \multicolumn{1}{l}{Augmented data}                   & \multicolumn{1}{l}{}                                 & ~                                                    \\ 
\cline{2-14}
                 & MLP                                                                  & 0.6630                                               & -                                                   & -                                                   & -                                                   & -                                                   & -                                                   & 0.7174                                               & -                                                     & -                                                    & -                                                    & -                                                    & -                                                    \\ 
\cline{2-14}
                 & LSTM                                                                 & 0.6685                                               & -                                                   & -                                                   & -                                                   & -                                                   & -                                                   & 0.7228                                               & -                                                     & -                                                    & -                                                    & -                                                    & -                                                    \\ 
\cline{2-14}
                 & GMRNN                                                                & 0.6923                                               & 0.7014                                              & 0.6988                                              & 0.7001                                              & 0.9709                                              & 0.8162                                              & 0.7479                                               & 0.7473                                                & 0.7449                                               & 0.7461                                               & 0.9735                                               & 0.8462                                               \\ 
\cline{2-14}
                 & IndRNN                                                               & 0.6923                                               & 0.7014                                              & 0.6988                                              & 0.7001                                              & 0.9709                                              & 0.8162                                              & 0.7579                                               & 0.7473                                                & 0.745                                                & 0.7461                                               & 0.9735                                               & 0.8462                                               \\ 
\cline{2-14}
                 & Transformer                                                          & 0.7423                                               & 0.7473                                              & 0.7449                                              & 0.7461                                              & -                                                   & -                                                   & 0.7689                                               & 0.7650                                                & 0.7571                                               & 0.7610                                               & -                                                    & -                                                    \\ 
\cline{2-14}
                 & Transformer+ Binary encoding                                         & \textbf{0.7712}                                               & \textbf{0.7714}                                              & \textbf{0.7799}                                              & \textbf{0.7756}                                              & -                                                   & -                                                   & \textbf{0.7908}                                               & \textbf{0.8000}                                                & \textbf{0.812}                                                & \textbf{0.8059}                                               & -                                                    & -                                                    \\ 
\hline
Image            & AlexNet                                                              & 0.3533                                               & -                                                   & -                                                   & -                                                   & -                                                   & -                                                   & 0.3750                                               & -                                                     & -                                                    & -                                                    & -                                                    & -                                                    \\ 
\cline{2-14}
                 & VGG16                                                                & 0.6576                                              & -                                                   & -                                                   & -                                                   & -                                                   & -                                                   & 0.7173                                               & -                                                     & -                                                    & -                                                    & -                                                    & -                                                    \\ 
\cline{2-14}
                 & VGG19                                                                & 0.5652                                               & -                                                   & -                                                   & -                                                   & -                                                   & -                                                   & 0.6304                                               & -                                                     & -                                                    & -                                                    & -                                                    & -                                                    \\ 
\cline{2-14}
                 & InceptionV3                                                          & 0.5109                                               & -                                                   & -                                                   & -                                                   & -                                                   & -                                                   & 0.5609                                               & -                                                     & -                                                    & -                                                    & -                                                    & -                                                    \\ 
\cline{2-14}
                 & ResNet50                                                             & 0.3370                                               & -                                                   & -                                                   & -                                                   & -                                                   & -                                                   & 0.3370                                               & -                                                     & -                                                    & -                                                    & -                                                    & -                                                    \\ 
\cline{2-14}
                 & Modified VGG16                                                       & \begin{tabular}[c]{@{}l@{}}0.5321 \\ \\\end{tabular} & \begin{tabular}[c]{@{}l@{}}0.6119\\ \\\end{tabular} & \begin{tabular}[c]{@{}l@{}}0.4469\\ \\\end{tabular} & 0.5162                                              & \begin{tabular}[c]{@{}l@{}}0.8261\\ \\\end{tabular} & \begin{tabular}[c]{@{}l@{}}0.8732\\ \\\end{tabular} & \begin{tabular}[c]{@{}l@{}}0.5870\\ \\\end{tabular}  & \begin{tabular}[c]{@{}l@{}}0.5914~ \\ \\\end{tabular} & \begin{tabular}[c]{@{}l@{}}0.5914\\ \\\end{tabular}  & \begin{tabular}[c]{@{}l@{}}0.5914\\ \\\end{tabular}  & \begin{tabular}[c]{@{}l@{}}0.7446\\ \\\end{tabular}  & \begin{tabular}[c]{@{}l@{}}0.9094\\ \\\end{tabular}  \\ 
\cline{2-14}
                 & Modified VGG19                                                       & \begin{tabular}[c]{@{}l@{}}0.5129 \\ \\\end{tabular} & \begin{tabular}[c]{@{}l@{}}0.6204\\ \\\end{tabular} & 0.3283                                              & \begin{tabular}[c]{@{}l@{}}0.4292\\ \\\end{tabular} & \begin{tabular}[c]{@{}l@{}}0.7609\\ \\\end{tabular} & \begin{tabular}[c]{@{}l@{}}0.8279\\ \\\end{tabular} & \begin{tabular}[c]{@{}l@{}}0.6087\\ \\\end{tabular}  & \begin{tabular}[c]{@{}l@{}}0.6624\\ \\\end{tabular}   & \begin{tabular}[c]{@{}l@{}}0.4876\\ \\\end{tabular}  & \begin{tabular}[c]{@{}l@{}}0.5609\\ \\\end{tabular}  & 0.8478                                               & \begin{tabular}[c]{@{}l@{}}0.8986\\ \\\end{tabular}  \\ 
\cline{2-14}
                 & Modified ResNet50                                                    & 0.6253                                               & 0.6371                                              & 0.6300                                              & 0.6335                                              & 0.7663                                              & 0.9293                                              & 0.7880                                               & 0.6995                                                & 0.6917                                               & 0.6955                                               & 0.7880                                               & 0.9547                                               \\ 
\cline{2-14}
                 & Modified InceptionResNetV2                                           & 0.7394                                               & 0.7486                                              & 0.7438                                              & 0.7462                                              & 0.8587                                              & 0.9656                                              & 0.7717                                               & 0.7806                                                & 0.7736                                               & 0.7771                                               & 0.9185                                               & 0.9783                                               \\
                 & Modified MobileNet                                                   & 0.7126                                               & 0.7237                                              & 0.7157                                              & 0.7197                                              & 0.8424                                              & 0.9638                                              & 0.7500                                               & 0.7653                                                & 0.7586                                               & 0.7619                                               & 0.8315                                               & 0.9565                                               \\ 
\cline{2-14}
                 & Modified DenseNet169                                                 & 0.7276                                               & 0.7262                                              & 0.7190                                              & 0.7226                                              & 0.8533                                              & 0.9601                                              & \begin{tabular}[c]{@{}l@{}}0.7174 \\ \\\end{tabular} & \begin{tabular}[c]{@{}l@{}}0.7247 \\ \\\end{tabular}  & \begin{tabular}[c]{@{}l@{}}0.7038 \\ \\\end{tabular} & 0.7140                                               & \begin{tabular}[c]{@{}l@{}}0.7880 \\ \\\end{tabular} & \begin{tabular}[c]{@{}l@{}}0.9511\\ \\\end{tabular}  \\ 
\cline{2-14}
                 & Modified EfficientNetB4                                              & 0.78893                                              & 0.7955                                              & 0.7955                                              & 0.7955                                              & 0.8859                                              & 0.9710                                              & \begin{tabular}[c]{@{}l@{}}0.8152 \\ \\\end{tabular} & \begin{tabular}[c]{@{}l@{}}0.8278 \\ \\\end{tabular}  & \begin{tabular}[c]{@{}l@{}}0.8224 \\ \\\end{tabular} & \begin{tabular}[c]{@{}l@{}}0.8251\\ \\\end{tabular}  & \begin{tabular}[c]{@{}l@{}}0.8913 \\ \\\end{tabular} & \begin{tabular}[c]{@{}l@{}}0.9710\\ \\\end{tabular}  \\ 
\cline{2-14}
                 & \begin{tabular}[c]{@{}l@{}}Modified EfficientNetV2M \\~\end{tabular} & 0.6323                                               & 0.6382                                              & 0.6305                                              & 0.6343                                              & 0.8043                                              & 0.9330                                              & 0.7446                                               & \begin{tabular}[c]{@{}l@{}}0.7662 \\ \\\end{tabular}  & 0.7514                                               & \begin{tabular}[c]{@{}l@{}}0.7587 \\ \\\end{tabular} & \begin{tabular}[c]{@{}l@{}}0.8315 \\ \\\end{tabular} & \begin{tabular}[c]{@{}l@{}}0.9511\\ \\\end{tabular}  \\ 
\cline{2-14}
                 & Vit                                                                  & 0.799                                                & 0.801                                               & \multicolumn{1}{l}{0.8045}                       & 0.8027                                              & -                                                   & -                                                   & 0.8283                                               & 0.8352                                                & 0.8312                                               & \begin{tabular}[c]{@{}l@{}}0.8331 \\~\end{tabular}   & -                                                    & -                                                    \\ 
\cline{2-14}
                 & Vit+ Wavelet                                                         & \textbf{0.8123}                                               & \textbf{0.8199}                                              & \textbf{0.8212}                                              & \textbf{0.8205}                                              & -                                                   & -                                                   & \textbf{0.8355}                                               & \textbf{0.8412}                                                & \textbf{0.844}                                                & \begin{tabular}[c]{@{}l@{}}\textbf{0.8425} \\~\end{tabular}   & -                                                    & -                                                    \\ 
\hline
Image + Location & AlexNet + MLP                                                        & 0.5543                                               & -                                                   & -                                                   & -                                                   & -                                                   & -                                                   & 0.6141                                               & -                                                     & -                                                    & -                                                    & -                                                    & -                                                    \\ 
\cline{2-14}
                 & VGG16 + MLP                                                          & 0.7717                                               & -                                                   & -                                                   & -                                                   & -                                                   & -                                                   & 0.78                                                 & -                                                     & -                                                    & -                                                    & -                                                    & -                                                    \\ 
\cline{2-14}
                 & VGG19 + MLP                                                          & 0.6250                                               & -                                                   & -                                                   & -                                                   & -                                                   & -                                                   & 0.7228                                               & -                                                     & -                                                    & -                                                    & -                                                    & -                                                    \\ 
\cline{2-14}
                 & InceptionV3 + MLP                                                    & 0.6141                                               & -                                                   & -                                                   & -                                                   & -                                                   & -                                                   & 0.711                                                & -                                                     & -                                                    & -                                                    & -                                                    & -                                                    \\ 
\cline{2-14}
                 & ResNet50 + MLP                                                       & 0.6304                                               & -                                                   & -                                                   & -                                                   & -                                                   & -                                                   & 0.6685                                               & -                                                     & -                                                    & -                                                    & -                                                    & -                                                    \\ 
\cline{2-14}
                 & AlexNet + LSTM                                                       & 0.5815                                               & -                                                   & -                                                   & -                                                   & -                                                   & -                                                   & 0.6685                                               & -                                                     & -                                                    & -                                                    & -                                                    & -                                                    \\ 
\cline{2-14}
                 & VGG16 + LSTM                                                         & 0.7283                                               & -                                                   & -                                                   & -                                                   & -                                                   & -                                                   & 0.7935                                               & -                                                     & -                                                    & -                                                    & -                                                    & -                                                    \\ 
\cline{2-14}
                 & VGG19 + LSTM                                                         & 0.71200                                              & -                                                   & -                                                   & -                                                   & -                                                   & -                                                   & 0.7663                                               & -                                                     & -                                                    & -                                                    & -                                                    & -                                                    \\ 
\cline{2-14}
                 & InceptionV3 + LSTM                                                   & 0.6467                                               & -                                                   & -                                                   & -                                                   & -                                                   & -                                                   & 0.692                                                & -                                                     & -                                                    & -                                                    & -                                                    & -                                                    \\ 
\cline{2-14}
                 & ResNet50 + LSTM                                                      & 0.3370                                               & -                                                   & -                                                   & -                                                   & -                                                   & -                                                   & 0.3479                                               & -                                                     & -                                                    & -                                                    & -                                                    & -                                                    \\ 
\cline{2-14}
                 & Xception+ GMRNN                                                      & 0.7877                                               & 0.7882                                              & 0.7715                                              & 0.7797                                              & 0.9662                                              & 0.9334                                              & 0.8189                                               & 0.8159                                                & \textbf{0.8469}                                               & 0.8311                                               & 0.8965                                               & 0.8944                                               \\ 
\cline{2-14}
                 & Vit+ Transformer                                                     & 0.7876                                               & 0.7881                                              & 0.7714                                              & 0.7797                                              & ~-                                                  & ~-                                                  & 0.8188                                               & 0.8158                                                & 0.8468                                               & 0.8310                                               & -                                                    & -                                                    \\ 
\cline{2-14}
                 & Vit+ Wavelet+ Transformer                                            & \textbf{0.8007}                                               & \textbf{0.8156}                                              & \textbf{0.8145}                                              & \textbf{0.8150}                                              & ~-                                                  & ~-                                                  & \textbf{0.8354}                                               & \textbf{0.8354}                                                & \multicolumn{1}{l}{0.8354}                           & \textbf{0.8354}                                               & -                                                    & -                                                    \\
\hline
\end{tabular}}
\caption{Performance comparison of models for four-class wound classification (D vs. P vs. S vs. V) on the AZH dataset using the original body map (484 locations). Results are reported for three input types (Location, Image, Image+Location) across original and augmented data. Bold values indicate the highest accuracy achieved within each input type.}
\label{Table1}
\end{table}

Table \ref{Table2} compares the results of the four wound class classifications (D vs. P vs. S vs. V) on the AZH dataset with a simplified body map containing 323 locations. On the Location data, the MLP model achieved an accuracy of 0.7174, and the LSTM achieved an accuracy of 0.7228, which had the weakest results on this dataset in the Original data mode. The two IndRNN and GMRNN models also achieved equal accuracy on the input dataset in the Original data mode, while the GMRNN model performed better in the Augmented data mode. The Transformer model was considered to have two input modes: numeric and binary input. In the numeric input mode, the accuracy of the Transformer model varied from 0.7553 and 0.7782 for the original and augmented data, respectively. In the Location input, the best result was obtained by Transformer + Binary encoding. In the Image + Location dataset in the Original data mode, the VGG models combined with MLP and LSTM achieved a maximum accuracy of 0.7935. These models achieved a maximum accuracy of 0.8152 in the Augmented data mode. The highest accuracy in both the Original and Augmented data modes was achieved by the Vit+ Wavelet +Transformer model. The Xception+ GMRNN model achieved an accuracy of 0.7991 in the Original data.

\begin{figure*}
\centering
\subfloat[ Location]{\includegraphics[width=1\textwidth]{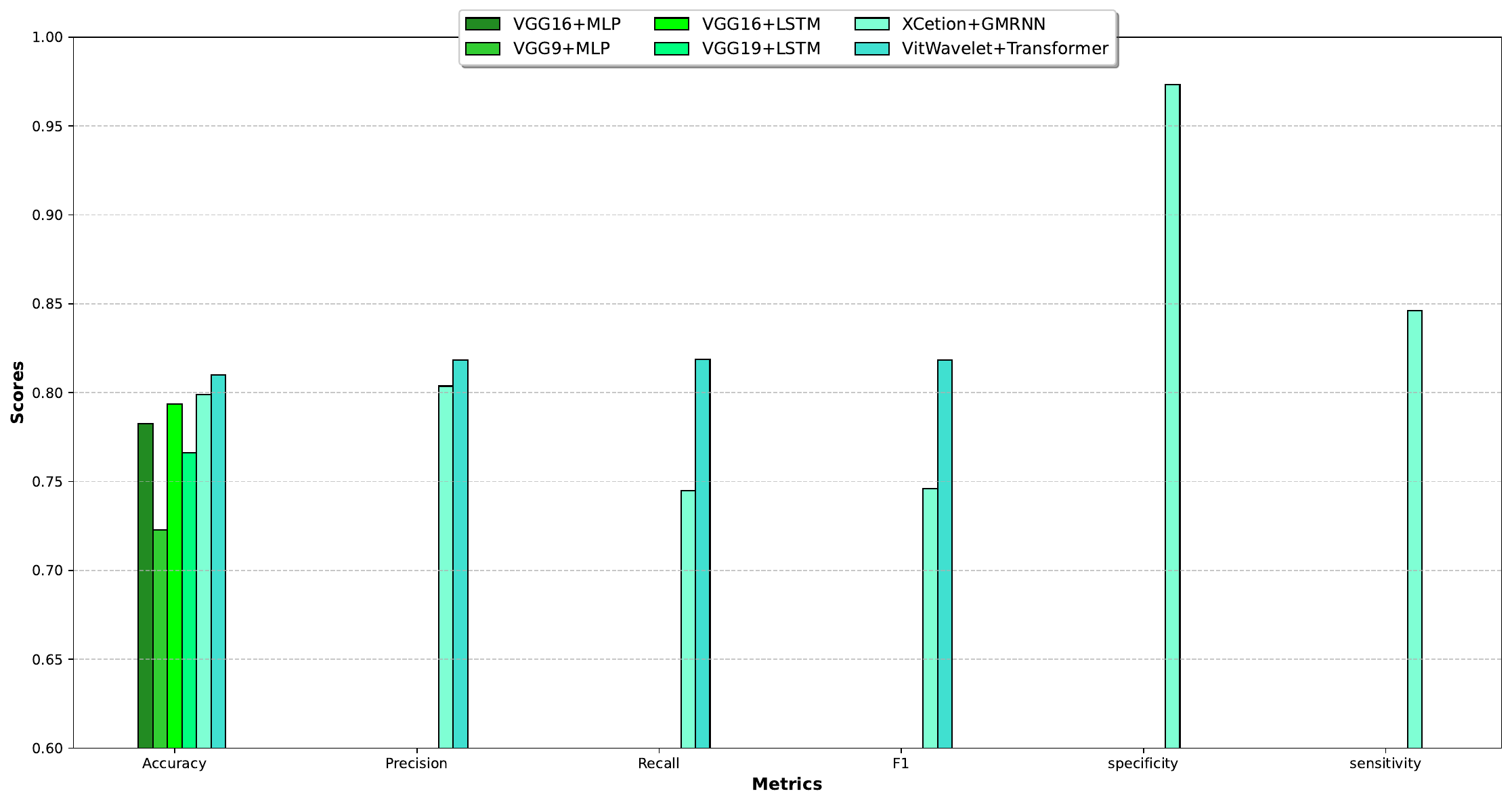}}\\
\subfloat[Image + Location]{\includegraphics[width=1\textwidth]{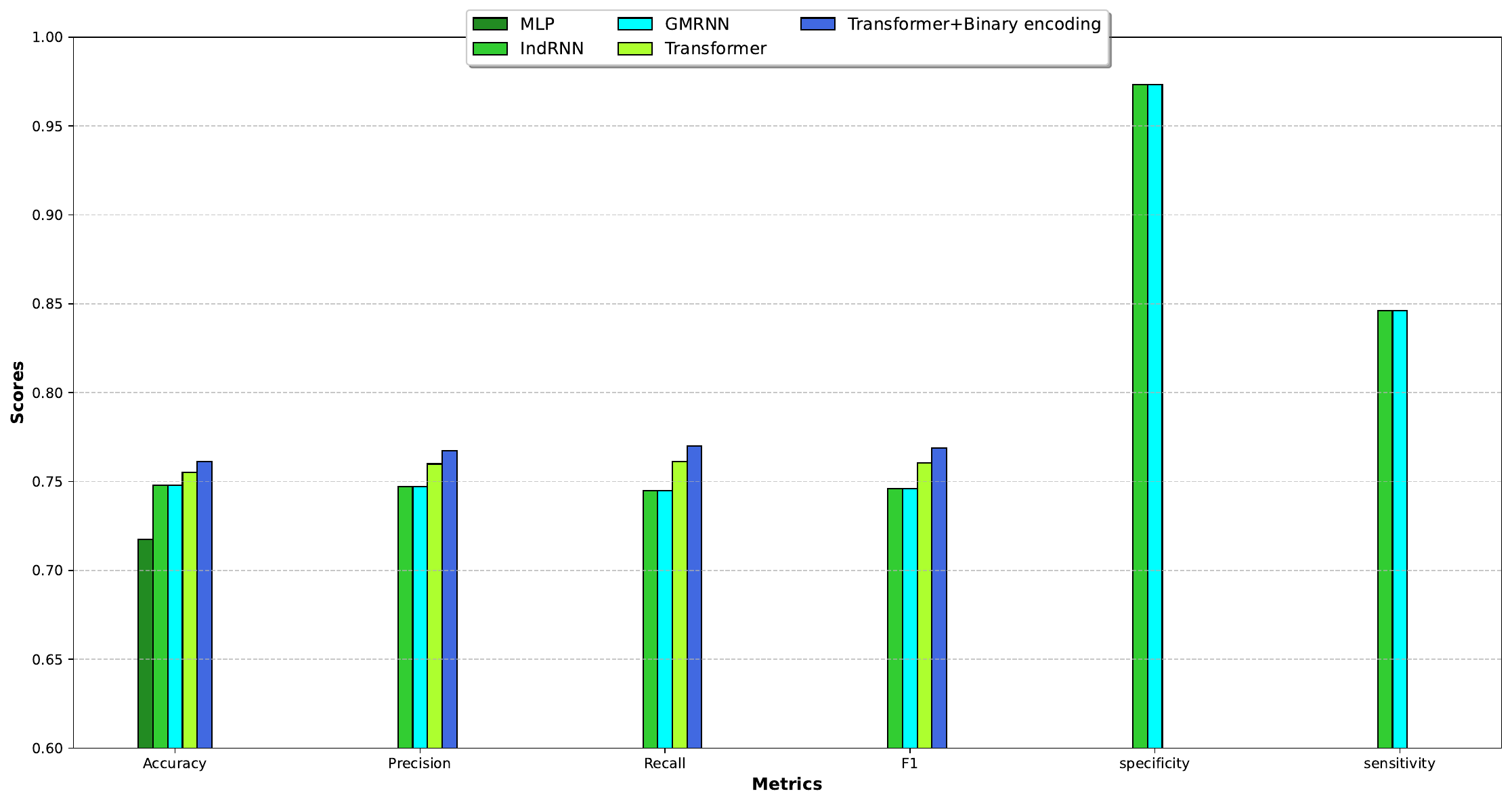}}\\
\caption{Bar plot for four wound class classification (D vs. P vs. S vs. V) on AZH dataset(Original Data).}
\label{fig:fig5}
\end{figure*}

Figure \ref{fig:fig5} shows the performance comparison of models for four-class wound classification (D vs. P vs. S vs. V) using the AZH dataset with a simplified body map of 323 locations. In subplot (a), which presents results for Location input, the ViT+Wavelet+Transformer model recorded the highest values across accuracy, precision, recall, and F1-score. The Xception+GMRNN model also performed similarly, with much greater specificity and sensitivity. Baseline models such as VGG16+MLP and VGG19+LSTM performed poorly in all metrics tested. In subplot (b), the combination of Location and Image input with Transformer+Binary encoding produced the best results in all important metrics, highlighting the merits of binary encoding. Against MLP and IndRNN, the Transformer-based methods significantly outperformed them in recall and F1-score. These results reinforce the case for using hybrid and transformer-based models, especially when coupled with structured input encoding.

\begin{figure*}
\centering
\subfloat[ Location]{\includegraphics[width=1\textwidth]{T2locationOrginal.pdf}}\\
\subfloat[Image + Location]{\includegraphics[width=1\textwidth]{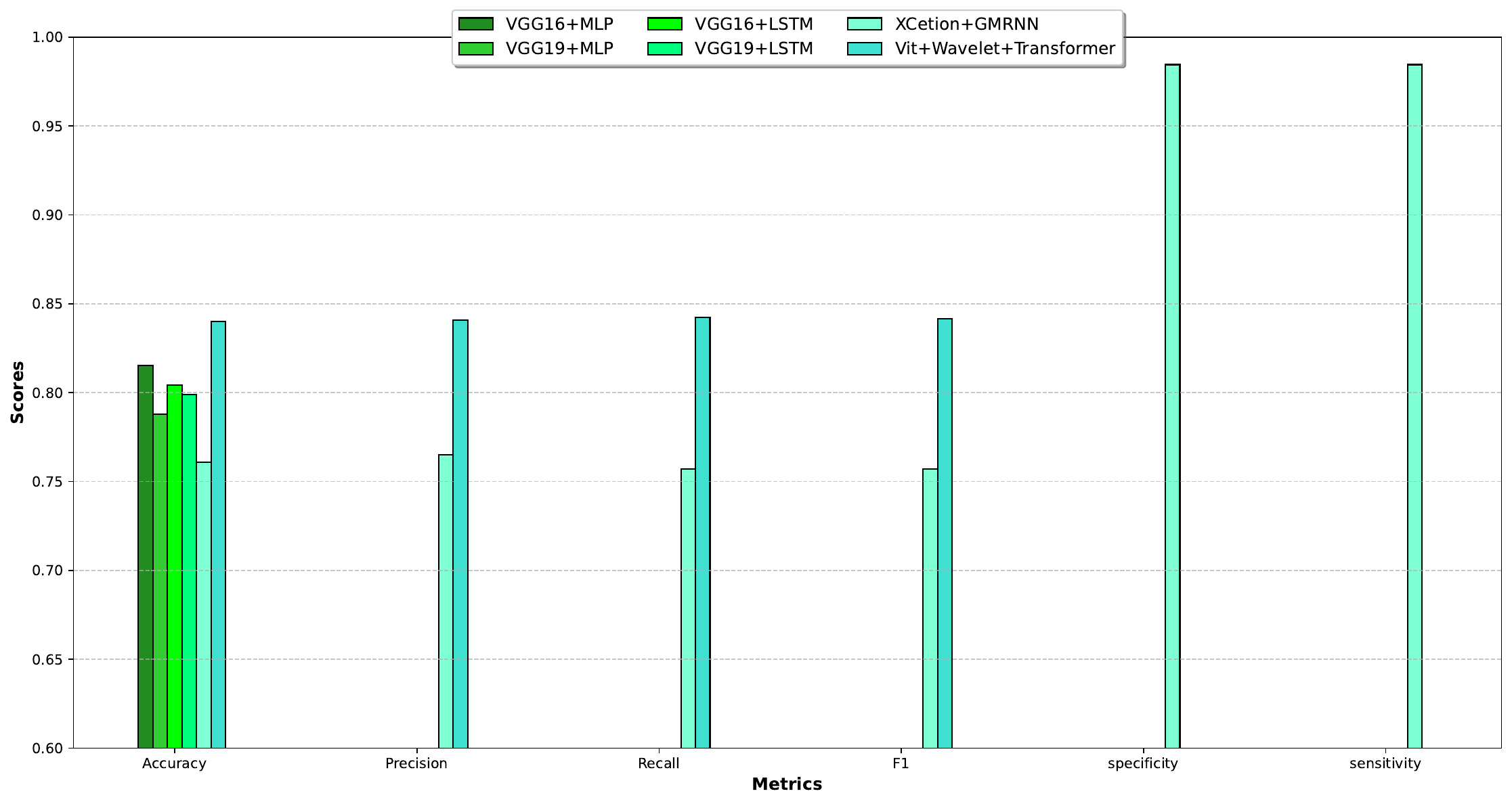}}\\
\caption{Bar plot for four wound class classification (D vs. P vs. S vs. V) on AZH dataset(Augmented Data).}
\label{fig:fig6}
\end{figure*}

Figure \ref{fig:fig6} shows that on the augmented AZH dataset, the ViT+Wavelet+Transformer model consistently outperformed others in both Location and Image+Location inputs, achieving the highest scores across all metrics, while Transformer+Binary encoding also showed strong results for Location-only input.

\begin{table}[]
\resizebox{\textwidth}{!}{
\begin{tabular}{|l|l|cccccc!{\vrule width 1pt}ccccc|c|}
\hline
\rowcolor{Gainsboro!60}
                                  &                              & \multicolumn{1}{c|}{Accuracy} & \multicolumn{1}{c|}{Precision} & \multicolumn{1}{c|}{Recall} & \multicolumn{1}{c|}{F1}     & \multicolumn{1}{c|}{specificity} & sensitivity & \multicolumn{1}{c|}{Accuracy} & \multicolumn{1}{c|}{Precision} & \multicolumn{1}{c|}{Recall} & \multicolumn{1}{c|}{F1}     & specificity & sensitivity \\ \hline
\multirow{7}{*}{Location}         &                              & \multicolumn{6}{c|}{Original Data}                                                                                                                                          & \multicolumn{5}{c|}{Augmented data}                                                                                                      &             \\ \cline{2-14} 
                                  & MLP                          & \multicolumn{1}{c|}{0.7174}   & \multicolumn{1}{c|}{-}         & \multicolumn{1}{c|}{-}      & \multicolumn{1}{c|}{-}      & \multicolumn{1}{c|}{-}           & -           & \multicolumn{1}{c|}{0.7446}   & \multicolumn{1}{c|}{-}         & \multicolumn{1}{c|}{-}      & \multicolumn{1}{c|}{-}      & -           & -           \\ \cline{2-14} 
                                  & LSTM                         & \multicolumn{1}{c|}{0.7228}   & \multicolumn{1}{c|}{-}         & \multicolumn{1}{c|}{-}      & \multicolumn{1}{c|}{-}      & \multicolumn{1}{c|}{-}           & -           & \multicolumn{1}{c|}{0.7337}   & \multicolumn{1}{c|}{-}         & \multicolumn{1}{c|}{-}      & \multicolumn{1}{c|}{-}      & -           & -           \\ \cline{2-14} 
                                  & IndRNN                       & \multicolumn{1}{c|}{0.7479}   & \multicolumn{1}{c|}{0.7473}    & \multicolumn{1}{c|}{0.7449} & \multicolumn{1}{c|}{0.7461} & \multicolumn{1}{c|}{0.9735}      & 0.8462      & \multicolumn{1}{c|}{0.7607}   & \multicolumn{1}{c|}{0.7650}    & \multicolumn{1}{c|}{0.7571} & \multicolumn{1}{c|}{0.7571} & 0.9846      & 0.8932      \\ \cline{2-14} 
                                  & GMRNN                        & \multicolumn{1}{c|}{0.7479}   & \multicolumn{1}{c|}{0.7473}    & \multicolumn{1}{c|}{0.7449} & \multicolumn{1}{c|}{0.7461} & \multicolumn{1}{c|}{0.9735}      & 0.8462      & \multicolumn{1}{c|}{0.7607}   & \multicolumn{1}{c|}{0.7650}    & \multicolumn{1}{c|}{0.7571} & \multicolumn{1}{c|}{0.7571} & 0.9846      & 0.8932      \\ \cline{2-14} 
                                  & Transformer                  & \multicolumn{1}{c|}{0.7553}   & \multicolumn{1}{c|}{0.7599}    & \multicolumn{1}{c|}{0.7612} & \multicolumn{1}{c|}{0.7605} & \multicolumn{1}{c|}{-}           & -           & \multicolumn{1}{c|}{0.7782}   & \multicolumn{1}{c|}{0.7923}    & \multicolumn{1}{c|}{0.7945}  & \multicolumn{1}{c|}{}       & -           & -           \\ \cline{2-14} 
                                  & Transformer+ Binary encoding & \multicolumn{1}{c|}{\textbf{0.7612}}   & \multicolumn{1}{c|}{\textbf{0.7675}}    & \multicolumn{1}{c|}{\textbf{0.77}}   & \multicolumn{1}{c|}{\textbf{0.7687}} & \multicolumn{1}{c|}{-}           & -           & \multicolumn{1}{c|}{\textbf{0.7899}}   & \multicolumn{1}{c|}{\textbf{0.8012}}    & \multicolumn{1}{c|}{\textbf{0.8023}} & \multicolumn{1}{c|}{}       & -           & -           \\ \hline
\multirow{6}{*}{Image + Location} & VGG16 + MLP                  & \multicolumn{1}{c|}{0.7826}   & \multicolumn{1}{c|}{-}         & \multicolumn{1}{c|}{-}      & \multicolumn{1}{c|}{-}      & \multicolumn{1}{c|}{-}           & -           & \multicolumn{1}{c|}{0.8152}   & \multicolumn{1}{c|}{-}         & \multicolumn{1}{c|}{-}      & \multicolumn{1}{c|}{-}      & -           & -           \\ \cline{2-14} 
                                  & VGG19 + MLP                  & \multicolumn{1}{c|}{0.7228}   & \multicolumn{1}{c|}{-}         & \multicolumn{1}{c|}{-}      & \multicolumn{1}{c|}{-}      & \multicolumn{1}{c|}{-}           & -           & \multicolumn{1}{c|}{0.7880}   & \multicolumn{1}{c|}{-}         & \multicolumn{1}{c|}{-}      & \multicolumn{1}{c|}{-}      & -           & -           \\ \cline{2-14} 
                                  & VGG16 +   LSTM               & \multicolumn{1}{c|}{0.7935}   & \multicolumn{1}{c|}{-}         & \multicolumn{1}{c|}{-}      & \multicolumn{1}{c|}{-}      & \multicolumn{1}{c|}{-}           & -           & \multicolumn{1}{c|}{0.8043}   & \multicolumn{1}{c|}{-}         & \multicolumn{1}{c|}{-}      & \multicolumn{1}{c|}{-}      & -           & -           \\ \cline{2-14} 
                                  & VGG19 +   LSTM               & \multicolumn{1}{c|}{0.7663}   & \multicolumn{1}{c|}{-}         & \multicolumn{1}{c|}{-}      & \multicolumn{1}{c|}{-}      & \multicolumn{1}{c|}{-}           & -           & \multicolumn{1}{c|}{0.7989}   & \multicolumn{1}{c|}{-}         & \multicolumn{1}{c|}{-}      & \multicolumn{1}{c|}{-}      & -           & -           \\ \cline{2-14} 
                                  & XCetion+ GMRNN               & \multicolumn{1}{c|}{0.7991}   & \multicolumn{1}{c|}{0.8037}    & \multicolumn{1}{c|}{0.7449} & \multicolumn{1}{c|}{0.7461} & \multicolumn{1}{c|}{0.9735}      & 0.8462      & \multicolumn{1}{c|}{0.7607}   & \multicolumn{1}{c|}{0.7650}    & \multicolumn{1}{c|}{0.7571} & \multicolumn{1}{c|}{0.7571} & 0.9846      & 0.9846      \\ \cline{2-14} 
                                  & Vit+   Wavelet + Transformer & \multicolumn{1}{c|}{\textbf{0.810}}    & \multicolumn{1}{c|}{\textbf{0.8183}}    & \multicolumn{1}{c|}{\textbf{0.8189}} & \multicolumn{1}{c|}{\textbf{0.8185}} & \multicolumn{1}{c|}{-}           & -           & \multicolumn{1}{c|}{\textbf{0.8399}}   & \multicolumn{1}{c|}{\textbf{0.8409}}    & \multicolumn{1}{c|}{\textbf{0.8425}} & \multicolumn{1}{c|}{\textbf{0.8416}} & -           & -           \\ \hline
\end{tabular}}
\caption{Model performance on four-class wound classification (D vs. P vs. S vs. V) using a simplified body map (323 locations) under original and augmented data. Comparison includes models based on Location and Image+Location inputs. Bold indicates the highest accuracy in each setting.}
\label{Table2}
\end{table}

A classification between all classes was performed on the AZH dataset. Table \ref{Table3} shows the results of this six-class classification (BG vs. N vs. D vs. P vs. S vs. V). On the Location input, the Transformer+ Binary encoding model achieved an accuracy of 0.7398; on the Image input, the Vit+ Wavelet model achieved an accuracy of 0.8401; and on the Image+ Location input, the Vit+ wavelet+ Transformer model achieved an accuracy of 0.8644, which was the highest accuracy recorded in the six-class classification.

\begin{figure*}
\centering
\subfloat[ Location]{\includegraphics[width=0.75\textwidth]{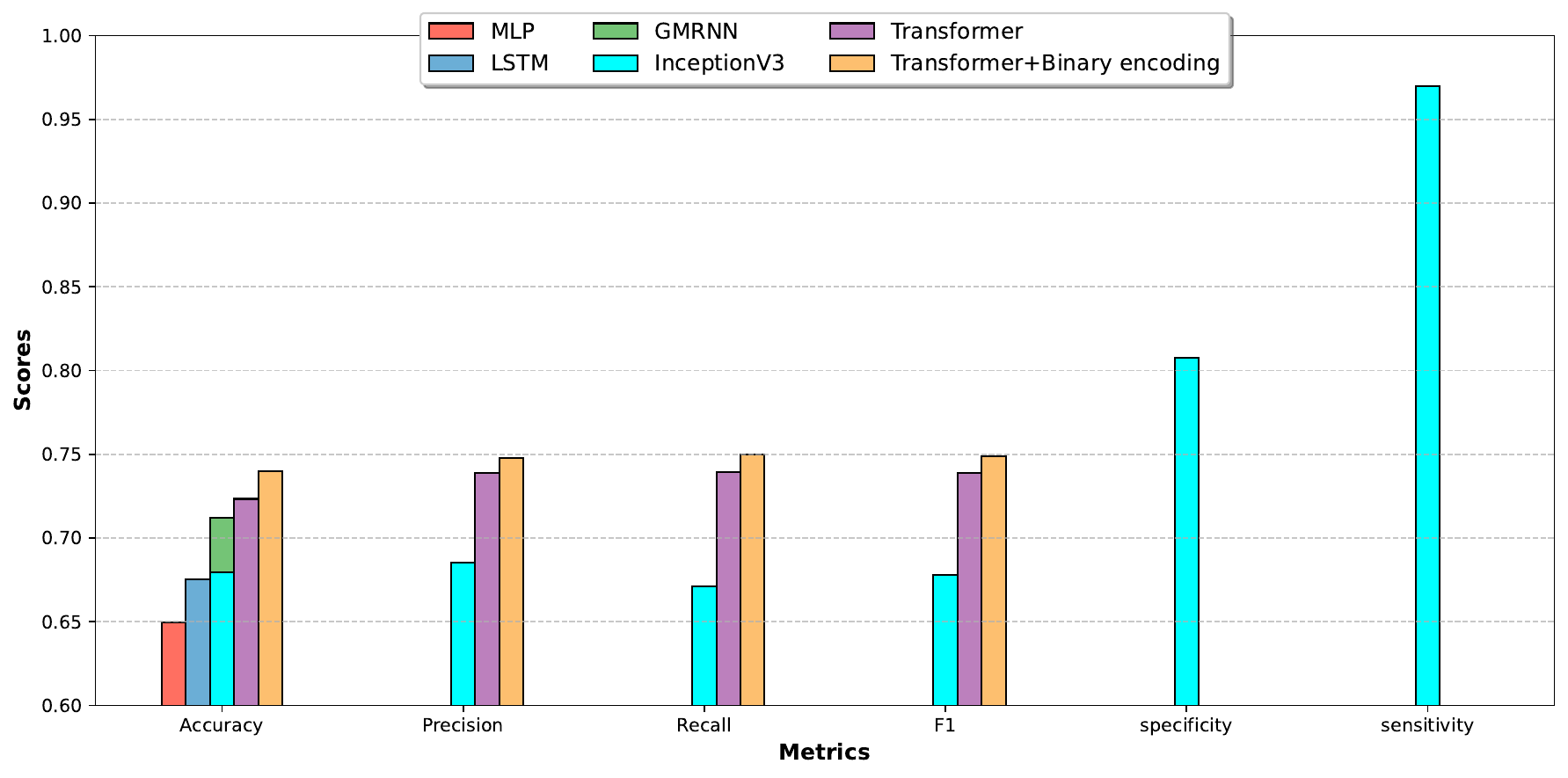}}\\
\subfloat[Image]{\includegraphics[width=0.75\textwidth]{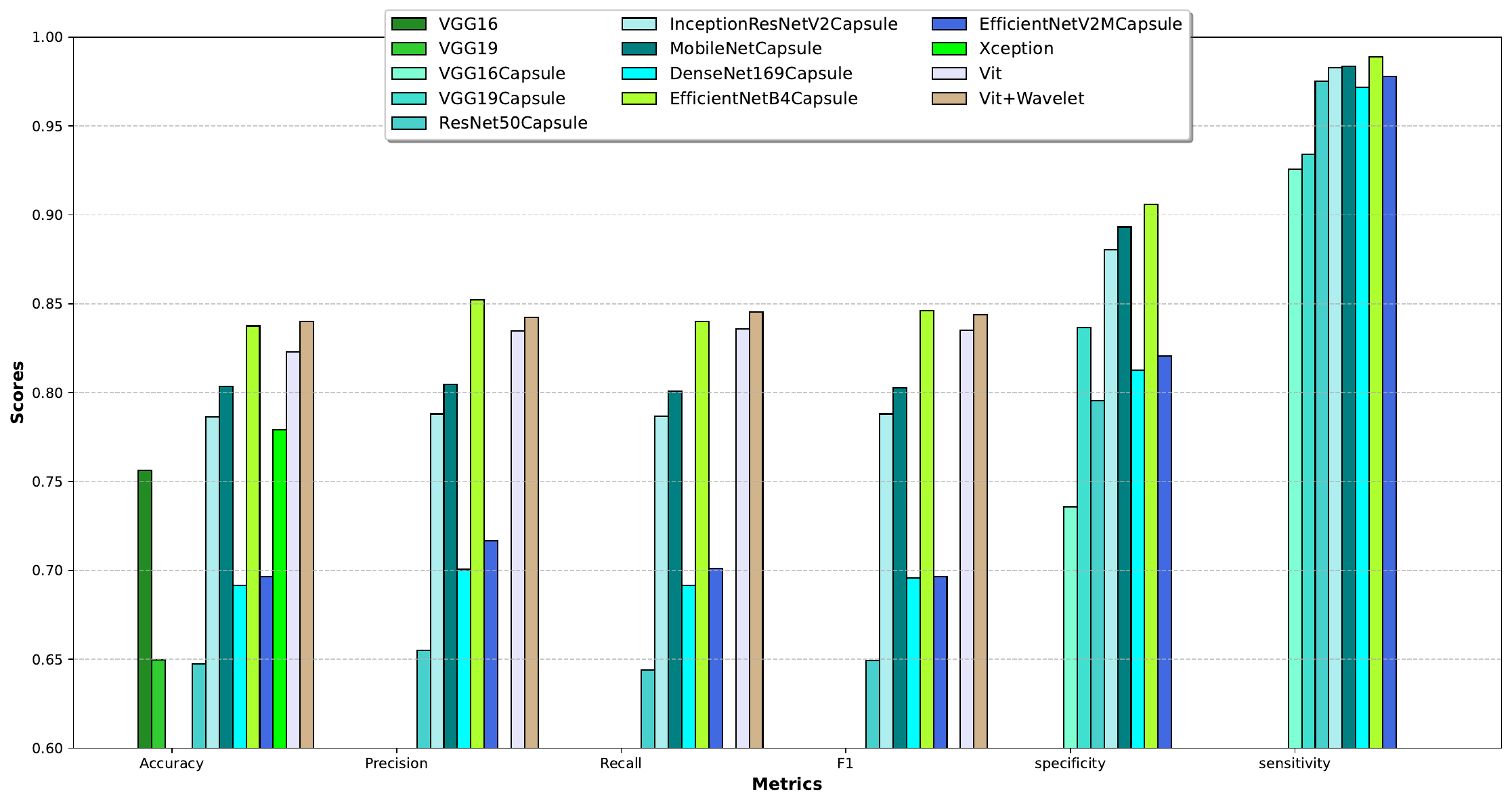}}\\
\subfloat[Image+Location]{\includegraphics[width=0.75\textwidth]{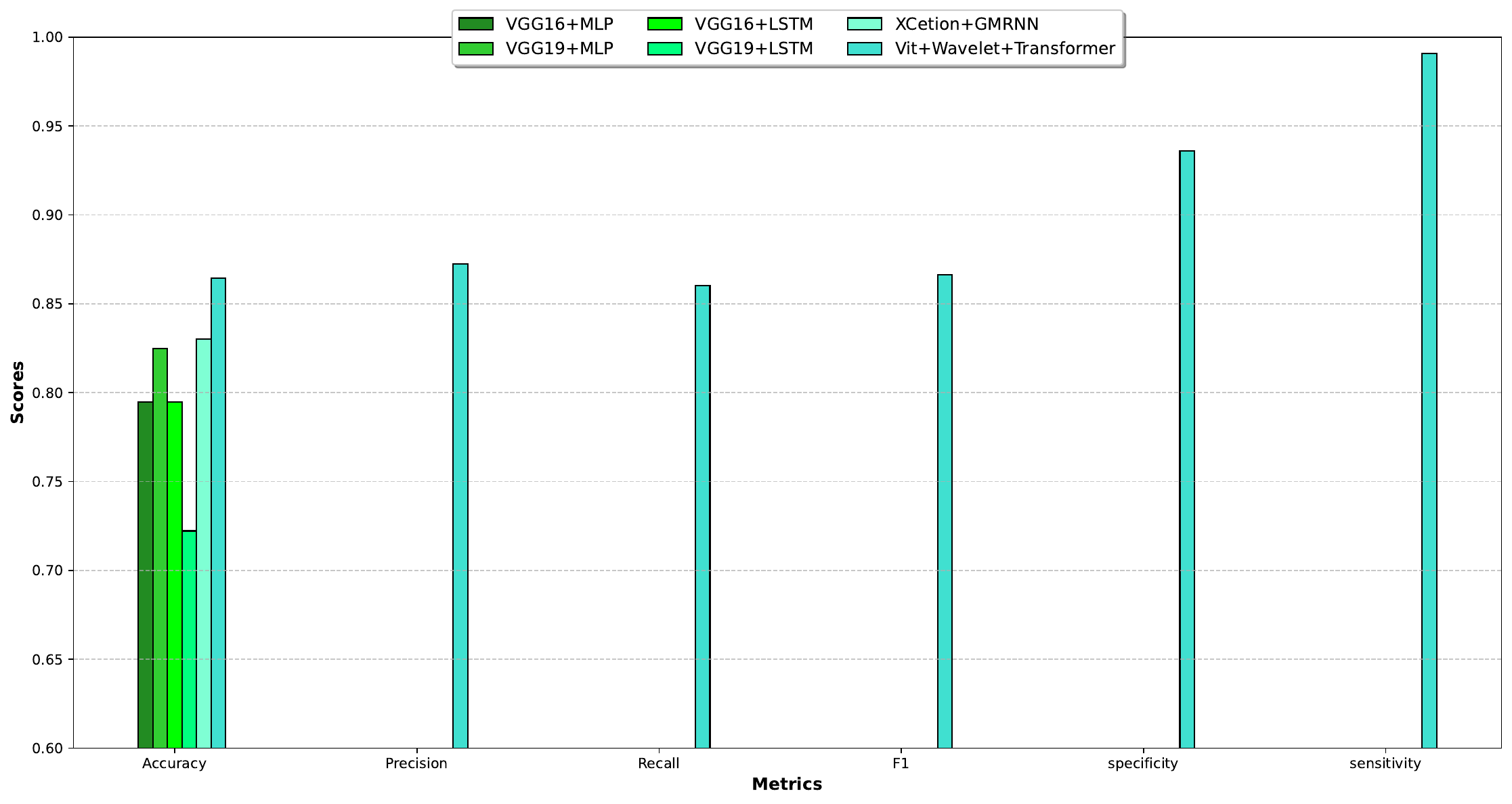}}\\
\caption{Bar plot for Six wound class classification (BG vs. N vs. D vs. P vs. S vs. V) on AZH dataset(Original Data).}
\label{fig:fig7}
\end{figure*}

Figure \ref{fig:fig7} illustrates the performance of models in the six-class wound classification task (BG vs. N vs. D vs. P vs. S vs. V) on the AZH dataset using Original Data. In subplot (a), which uses Location input, Transformer+Binary encoding outperformed all other models in terms of accuracy, precision, and recall. In subplot (b), based on Image input, the ViT+Wavelet model achieved the highest values across all metrics, closely followed by EfficientNetB4 and MobileNet variants. Capsule-based models also showed competitive performance in specificity and sensitivity. In subplot (c), which combines Image and Location inputs, ViT+Wavelet+Transformer achieved the best performance on all metrics, confirming its robustness in multi-class wound classification. These visualizations reinforce the advantage of combining transformer architectures with wavelet encoding, especially when both visual and positional information are used.

\begin{table}
\centering
\resizebox{\textwidth}{!}{
\begin{tblr}{
  vlines,
  hline{1-2,8,21-22,27} = {-}{},
  hline{3-7,9-20,23-26} = {2-8}{},
  row{1,2} = {Gainsboro!60},
}
Input             & Model                                  & Accuracy          & Precision & Recall            & F1                & specificity & sensitivity       \\
Location          & MLP                                    & 0.6496            & ~         & ~                 & ~                 & ~           & ~                 \\
                  & LSTM                                   & 0.6752            & ~         & ~                 & ~                 & ~           & ~                 \\
                  & IndRNN                                 & 0.6795            & 0.6852    & 0.6713            & 0.6781            & 0.8077      & 0.9701            \\
                  & GMRNN                                  & 0.712             & -         & -                 & -                 & -           & -                 \\
                  & Transformer                            & 0.7234            & 0.7390    & 0.7393            & ~                 & -           & -                 \\
                  & Transformer+
Binary encoding                     & \textbf{0.7398}            & \textbf{0.7478}    & \textbf{0.7501}            & ~                 & -           & -                 \\
Image             & VGG16                                  & 0.7564            & ~         & ~                 & ~                 & ~           & ~                 \\
                  & VGG19                                  & 0.6496            & ~         & ~                 & ~                 & ~           & ~                 \\
                  & Modified
  VGG16                       & 0.5433            & 0.5408    & 0.5263            & 0.5334            & 0.7357      & 0.9256            \\
                  & Modified
  VGG19                       & 0.5692            & 0.5876    & 0.5479            & 0.5666            & 0.8367      & 0.9342            \\
                  & Modified
  ResNet50                    & 0.6473            & 0.6551    & 0.6440            & 0.6494            & 0.7956      & 0.9752            \\
                  & Modified
  InceptionResNetV2           & 0.7863            & 0.7881    & 0.7866            & 0.7881            & 0.8803      & 0.9829            \\
                  & Modified
  MobileNet                   & 0.8034~           & 0.8047    & 0.8009            & 0.8028            & 0.8932      & 0.9838            \\
                  & Modified
  DenseNet169                 & 0.6914            & 0.7005    & 0.6915            & 0.6959            & 0.8128      & 0.9720            \\
                  & Modified
  EfficientNetB4              & 0.8376            & \textbf{0.8523}    & 0.8402            & \textbf{0.8462}            & 0.9060      & 0.9889            \\
                  & {
  Modified EfficientNetV2M
  \\~
  } & {
  0.6966\\~\\~} & 0.7167    & {
  0.7009\\~\\~} & {
  0.6966\\~\\~} & 0.8205      & {
  0.9778\\~\\~} \\
                  & Xception                               & 0.779             & -         & -                 & -                 & -           & -                 \\
                  & Vit                                    & 0.8231            & 0.8347    & 0.8359            & 0.8352            & -           & -                 \\
                  & Vit +
  Wavelet                        & \textbf{0.8401}            & 0.8423    & \textbf{0.8453}            & 0.8437            & -           & -                 \\
Image+
  location & VGG16+MLP                              & 0.7949            & -         & -                 & -                 & -           & -                 \\
                  & VGG19+MLP                              & 0.8248            & -         & -                 & -                 & -           & -                 \\
                  & VGG16+LSTM                             & 0.7949            & -         & -                 & -                 & -           & -                 \\
                  & VGG19+LSTM                             & 0.7222            & -         & -                 & -                 & -           & -                 \\
                  & Xception+ GMRNN                        & 0.83              & -         & -                 & -                 & -           & -                 \\
                  & Vit+
  wavelet+ Transformer            & \textbf{0.8644}            & \textbf{0.8723}    & \textbf{0.8602}            & \textbf{0.8662}            & \textbf{0.9360}      & \textbf{0.9909}            
\end{tblr}}
\caption{Results of six-class wound classification (BG vs. N vs. D vs. P vs. S vs. V) on the AZH dataset. Evaluated across Location, Image, and Image+Location inputs. The Vit+Wavelet+Transformer model achieves the highest accuracy overall.}
\label{Table3}
\end{table}

Four five-class classifications were performed on the AZH dataset(see Table \ref{Table4}). The classifications were (1) BG vs. N vs. D vs. P vs. V, (2) BG vs. N vs. D vs. S vs. V, (3) BG vs. N vs. D vs. P vs. S vs. V, and (4) BG vs. N vs. P vs. S vs. V. In the classification of the classes BG vs. N vs. D vs. P vs. V, the accuracy of the Transformer+ Binary Encoding model in the Location input was 0.7198, the Vit+ Wavelet model in the Image input was 0.8198, and the Vit+Wavelet+ Transformer model in the Image+Location input was 0.9042, which was the highest result recorded in this classification. In the classification of BG vs. N vs. D vs. S vs. V and Location input of the MLP model achieved an accuracy of 0.7500, on Image input, the Vit+ Wavelet model achieved an accuracy of 0.8012, and on Image+ Location input, the Xception+ GMRNN model achieved an accuracy of 0.9310. Also, in the BG vs. N vs. D vs. P vs. S classification, the highest accuracy was achieved on Location, Image, and Image+ Location input of the Transformer+ Binary Encoding, Vit+ Wavelet, and Xception+ GMRNN models, respectively. In BG vs. N vs. P vs. S vs. V, the highest accuracy was reported by the Xception+ GMRNN model with Image + Location input. This model achieved an accuracy of 0.8712.

\begin{table}[]
\resizebox{\textwidth}{!}{
\begin{tabular}{|l|l|cccc|}
\hline
\rowcolor{Gainsboro!60}
Input                            & Model                        & \multicolumn{1}{c|}{BG–N–D–P–V} & \multicolumn{1}{c|}{BG–N–D–S–V} & \multicolumn{1}{c|}{BG–N–D–P–S} & BG–N–P–S–V \\ \hline
                                 &                              & \multicolumn{4}{c|}{Accuracy}                                                                                    \\ \hline
\multirow{6}{*}{Location}        & MLP                          & \multicolumn{1}{c|}{0.6771}     & \multicolumn{1}{c|}{\textbf{0.7500}}     & \multicolumn{1}{c|}{0.5930}     & 0.6968     \\ \cline{2-6} 
                                 & LSTM                         & \multicolumn{1}{c|}{0.6875}     & \multicolumn{1}{c|}{0.7200}     & \multicolumn{1}{c|}{0.5930}     & 0.7181     \\ \cline{2-6} 
                                 & IndRNN                       & \multicolumn{1}{c|}{0.6924}     & \multicolumn{1}{c|}{0.7423}     & \multicolumn{1}{c|}{0.6234}     & 0.7250     \\ \cline{2-6} 
                                 & GMRNN                        & \multicolumn{1}{c|}{0.6920}     & \multicolumn{1}{c|}{0.7420}     & \multicolumn{1}{c|}{0.6230}     & 0.7050     \\ \cline{2-6} 
                                 & Transformer                  & \multicolumn{1}{c|}{0.7001}     & \multicolumn{1}{c|}{0.7243}     & \multicolumn{1}{c|}{0.6893}     & 0.7422     \\ \cline{2-6} 
                                 & Transformer+ Binary encoding & \multicolumn{1}{c|}{\textbf{0.7198}}     & \multicolumn{1}{c|}{0.7476}     & \multicolumn{1}{c|}{\textbf{0.6899}}     & \textbf{0.7443}     \\ \hline
\multirow{13}{*}{Image}          & VGG16                        & \multicolumn{1}{c|}{0.6979}     & \multicolumn{1}{c|}{0.7050}     & \multicolumn{1}{c|}{0.6453}     & 0.7553     \\ \cline{2-6} 
                                 & VGG19                        & \multicolumn{1}{c|}{0.7656}     & \multicolumn{1}{c|}{0.7450}     & \multicolumn{1}{c|}{0.6744}     & 0.7234     \\ \cline{2-6} 
                                 & Modified VGG16               & \multicolumn{1}{c|}{0.7774}     & \multicolumn{1}{c|}{0.7342}     & \multicolumn{1}{c|}{0.7331}     & 0.7510     \\ \cline{2-6} 
                                 & Modified VGG19               & \multicolumn{1}{c|}{0.7803}     & \multicolumn{1}{c|}{0.7325}     & \multicolumn{1}{c|}{0.7121}     & 0.7510     \\ \cline{2-6} 
                                 & Modified ResNet50            & \multicolumn{1}{c|}{0.7869}     & \multicolumn{1}{c|}{0.7423}     & \multicolumn{1}{c|}{0.7121}     & 0.7310     \\ \cline{2-6} 
                                 & Modified InceptionResNetV2   & \multicolumn{1}{c|}{0.7898}     & \multicolumn{1}{c|}{0.7553}     & \multicolumn{1}{c|}{0.7541}     & 0.7810     \\ \cline{2-6} 
                                 & Modified MobileNet           & \multicolumn{1}{c|}{0.7912}     & \multicolumn{1}{c|}{0.7543}     & \multicolumn{1}{c|}{0.7557}     & 0.7810     \\ \cline{2-6} 
                                 & Modified DenseNet169         & \multicolumn{1}{c|}{0.7941}     & \multicolumn{1}{c|}{0.7622}     & \multicolumn{1}{c|}{0.7234}     & 0.7610     \\ \cline{2-6} 
                                 & Modified EfficientNetB4      & \multicolumn{1}{c|}{0.7676}     & \multicolumn{1}{c|}{0.7733}     & \multicolumn{1}{c|}{0.7612}     & 0.7710     \\ \cline{2-6} 
                                 & Modified EfficientNetV2M     & \multicolumn{1}{c|}{0.8032}     & \multicolumn{1}{c|}{0.7712}     & \multicolumn{1}{c|}{0.7308}     & 0.7810     \\ \cline{2-6} 
                                 & Xception                     & \multicolumn{1}{c|}{0.7774}     & \multicolumn{1}{c|}{0.7723}     & \multicolumn{1}{c|}{0.7701}     & 0.7610     \\ \cline{2-6} 
                                 & Vit                          & \multicolumn{1}{c|}{0.8012}     & \multicolumn{1}{c|}{0.7888}     & \multicolumn{1}{c|}{0.7700}     & 0.7854     \\ \cline{2-6} 
                                 & Vit+ wavelet                 & \multicolumn{1}{c|}{\textbf{0.8198}}     & \multicolumn{1}{c|}{\textbf{0.8012}}     & \multicolumn{1}{c|}{\textbf{0.7829}}     & \textbf{0.7899}     \\ \hline
\multirow{6}{*}{Image+ location} & VGG16+MLP                    & \multicolumn{1}{c|}{0.8646}     & \multicolumn{1}{c|}{0.8500}     & \multicolumn{1}{c|}{0.8314}     & 0.8404     \\ \cline{2-6} 
                                 & VGG19+MLP                    & \multicolumn{1}{c|}{0.8542}     & \multicolumn{1}{c|}{0.8650}     & \multicolumn{1}{c|}{0.7733}     & 0.8617     \\ \cline{2-6} 
                                 & VGG16+LSTM                   & \multicolumn{1}{c|}{0.8438}     & \multicolumn{1}{c|}{0.9100}     & \multicolumn{1}{c|}{0.7733}     & 0.7713     \\ \cline{2-6} 
                                 & VGG19+LSTM                   & \multicolumn{1}{c|}{0.8438}     & \multicolumn{1}{c|}{0.9100}     & \multicolumn{1}{c|}{0.7733}     & 0.7713     \\ \cline{2-6} 
                                 & Xception+ GMRNN              & \multicolumn{1}{c|}{0.8885}     & \multicolumn{1}{c|}{\textbf{0.9310}}     & \multicolumn{1}{c|}{\textbf{0.8712}}     & \textbf{0.8712}     \\ \cline{2-6} 
                                 & Vit+ wavelet + Transformer   & \multicolumn{1}{c|}{\textbf{0.9042}}     & \multicolumn{1}{c|}{0.9250}     & \multicolumn{1}{c|}{0.7733}     & 0.8623     \\ \hline
\end{tabular}}
\caption{Accuracy results for four different five-class classification combinations on the AZH dataset. Models are evaluated under Location, Image, and Image+Location inputs. Bold values denote the highest accuracy achieved in each classification setting.}
\label{Table4}
\end{table}

The following four-class classification results for the classes (1) BG vs. N vs. D vs. V, (2) BG vs. N vs. P vs. V, (3) BG vs. N vs. S vs. V, (4) BG vs. N vs. D vs. P, (5) BG vs. N vs. D vs. S, and (6) BG vs. N vs. P vs. S are given in Table \ref{Table5}. The Transformer+ Binary encoding model achieved maximum accuracy in all classes. In the Image input, the Vit+ Wavelet model achieved the highest accuracy in  BG vs. N vs. D vs. V,  BG vs. N vs. P vs. V, (3) BG vs. N vs. S vs. V,  BG vs. N vs. D vs. S, and  BG vs. N vs. P vs. S. The EfficientNetB4+ CapsuleNet model achieved an accuracy of 0.8534, which was the highest accuracy in the Image input. In the Image + Location input, the Vit+ Wavelet+ Transformer model reported the highest accuracy, which was reported in the BG vs. N vs. D vs. V class. This model achieved an accuracy of 0.9709 in this class.


\begin{table}[]
\resizebox{\textwidth}{!}{
\begin{tabular}{|l|l|cccccc|}
\hline
\rowcolor{Gainsboro!60}
Input                             & Model                               & \multicolumn{1}{c|}{BG–N–D–V} & \multicolumn{1}{c|}{BG–N–P–V} & \multicolumn{1}{c|}{BG–N–S–V} & \multicolumn{1}{c|}{BG–N–D–P} & \multicolumn{1}{c|}{BG–N–D–S} & BG–N–P–S \\ \hline
                                  &                                     & \multicolumn{6}{c|}{Accuracy}                                                                                                                                            \\ \hline
\multirow{6}{*}{Location}         & MLP                                 & \multicolumn{1}{c|}{0.7658}   & \multicolumn{1}{c|}{0.7329}   & \multicolumn{1}{c|}{0.7727}   & \multicolumn{1}{c|}{0.6538}   & \multicolumn{1}{c|}{0.7174}   & 0.6904   \\ \cline{2-8} 
                                  & LSTM                                & \multicolumn{1}{c|}{0.7848}   & \multicolumn{1}{c|}{0.7603}   & \multicolumn{1}{c|}{0.8312}   & \multicolumn{1}{c|}{0.6462}   & \multicolumn{1}{c|}{0.7391}   & 0.6746   \\ \cline{2-8} 
                                  & IndRNN                              & \multicolumn{1}{c|}{0.8010}   & \multicolumn{1}{c|}{0.7621}   & \multicolumn{1}{c|}{0.8509}   & \multicolumn{1}{c|}{0.7144}   & \multicolumn{1}{c|}{0.7500}   & 0.7109   \\ \cline{2-8} 
                                  & GMRNN                               & \multicolumn{1}{c|}{0.8000}   & \multicolumn{1}{c|}{0.7601}   & \multicolumn{1}{c|}{0.8509}   & \multicolumn{1}{c|}{0.7101}   & \multicolumn{1}{c|}{0.7511}   & 0.7009   \\ \cline{2-8} 
                                  & Transformer                         & \multicolumn{1}{c|}{0.8001}   & \multicolumn{1}{c|}{0.7922}   & \multicolumn{1}{c|}{0.8713}   & \multicolumn{1}{c|}{0.7232}   & \multicolumn{1}{c|}{0.7543}   & 0.6922   \\ \cline{2-8} 
                                  & Transformer+ Binary encoding        & \multicolumn{1}{c|}{\textbf{0.8119}}   & \multicolumn{1}{c|}{\textbf{0.8000}}   & \multicolumn{1}{c|}{\textbf{0.8821}}   & \multicolumn{1}{c|}{\textbf{0.7432}}   & \multicolumn{1}{c|}{\textbf{0.7667}}   & \textbf{0.7332}   \\ \hline
\multirow{13}{*}{Image}           & VGG16                               & \multicolumn{1}{c|}{0.9367}   & \multicolumn{1}{c|}{0.8973}   & \multicolumn{1}{c|}{0.8766}   & \multicolumn{1}{c|}{0.8231}   & \multicolumn{1}{c|}{0.7754}   & 0.8333   \\ \cline{2-8} 
                                  & VGG19                               & \multicolumn{1}{c|}{0.8987}   & \multicolumn{1}{c|}{0.8699}   & \multicolumn{1}{c|}{0.8831}   & \multicolumn{1}{c|}{0.8000}   & \multicolumn{1}{c|}{0.8188}   & 0.8333   \\ \cline{2-8} 
                                  & Modified VGG16                      & \multicolumn{1}{c|}{0.8893}   & \multicolumn{1}{c|}{0.8898}   & \multicolumn{1}{c|}{0.9091}   & \multicolumn{1}{c|}{0.8400}   & \multicolumn{1}{c|}{0.8101}   & 0.8401   \\ \cline{2-8} 
                                  & Modified VGG19                      & \multicolumn{1}{c|}{0.9434}   & \multicolumn{1}{c|}{0.9121}   & \multicolumn{1}{c|}{0.9210}   & \multicolumn{1}{c|}{0.8401}   & \multicolumn{1}{c|}{0.8133}   & 0.8451   \\ \cline{2-8} 
                                  & Modified ResNet50                   & \multicolumn{1}{c|}{0.9555}   & \multicolumn{1}{c|}{0.9012}   & \multicolumn{1}{c|}{0.912}    & \multicolumn{1}{c|}{0.8494}   & \multicolumn{1}{c|}{0.8143}   & 0.8341   \\ \cline{2-8} 
                                  & Modified InceptionResNetV2          & \multicolumn{1}{c|}{0.9534}   & \multicolumn{1}{c|}{0.9090}   & \multicolumn{1}{c|}{0.9010}   & \multicolumn{1}{c|}{0.8344}   & \multicolumn{1}{c|}{0.8294}   & 0.8431   \\ \cline{2-8} 
                                  & Modified MobileNet                  & \multicolumn{1}{c|}{0.9512}   & \multicolumn{1}{c|}{0.9002}   & \multicolumn{1}{c|}{0.9192}   & \multicolumn{1}{c|}{0.8534}   & \multicolumn{1}{c|}{0.8342}   & 0.8531   \\ \cline{2-8} 
                                  & Modified DenseNet169                & \multicolumn{1}{c|}{0.9543}   & \multicolumn{1}{c|}{0.9142}   & \multicolumn{1}{c|}{0.9101}   & \multicolumn{1}{c|}{0.8405}   & \multicolumn{1}{c|}{0.8266}   & 0.8521   \\ \cline{2-8} 
                                  & Modified EfficientNetB4             & \multicolumn{1}{c|}{0.9691}   & \multicolumn{1}{c|}{0.9165}   & \multicolumn{1}{c|}{0.9102}   & \multicolumn{1}{c|}{0.8409}   & \multicolumn{1}{c|}{0.8523}   & 0.8241   \\ \cline{2-8} 
                                  & Modified EfficientNetV2M            & \multicolumn{1}{c|}{0.9453}   & \multicolumn{1}{c|}{0.9183}   & \multicolumn{1}{c|}{0.8949}   & \multicolumn{1}{c|}{\textbf{0.8509}}   & \multicolumn{1}{c|}{0.8534}   & 0.8631   \\ \cline{2-8} 
                                  & Xception                            & \multicolumn{1}{c|}{0.943}    & \multicolumn{1}{c|}{0.9111}   & \multicolumn{1}{c|}{0.9091}   & \multicolumn{1}{c|}{0.8422}   & \multicolumn{1}{c|}{0.8101}   & 0.8401   \\ \cline{2-8} 
                                  & Vit                                 & \multicolumn{1}{c|}{0.9578}   & \multicolumn{1}{c|}{0.9012}   & \multicolumn{1}{c|}{0.8978}   & \multicolumn{1}{c|}{0.8297}   & \multicolumn{1}{c|}{0.8398}   & 0.8500   \\ \cline{2-8} 
                                  & Vit+ Wavelet                        & \multicolumn{1}{c|}{\textbf{0.9767}}   & \multicolumn{1}{c|}{\textbf{0.9232}}   & \multicolumn{1}{c|}{\textbf{0.9209}}   & \multicolumn{1}{c|}{0.8393}   & \multicolumn{1}{c|}{\textbf{0.8555}}   & 0.8732   \\ \hline
\multirow{14}{*}{Image+ location} & VGG16+MLP                           & \multicolumn{1}{c|}{0.9430}   & \multicolumn{1}{c|}{0.9178}   & \multicolumn{1}{c|}{0.9416}   & \multicolumn{1}{c|}{0.8615}   & \multicolumn{1}{c|}{0.8615}   & 0.8571   \\ \cline{2-8} 
                                  & VGG19+MLP                           & \multicolumn{1}{c|}{0.9557}   & \multicolumn{1}{c|}{0.9178}   & \multicolumn{1}{c|}{0.9286}   & \multicolumn{1}{c|}{0.8692}   & \multicolumn{1}{c|}{0.9130}   & 0.8175   \\ \cline{2-8} 
                                  & VGG16+LSTM                          & \multicolumn{1}{c|}{0.8987}   & \multicolumn{1}{c|}{0.9247}   & \multicolumn{1}{c|}{0.9091}   & \multicolumn{1}{c|}{0.8615}   & \multicolumn{1}{c|}{0.8478}   & 0.8333   \\ \cline{2-8} 
                                  & VGG19+LSTM                          & \multicolumn{1}{c|}{0.9430}   & \multicolumn{1}{c|}{0.8904}   & \multicolumn{1}{c|}{0.8889}   & \multicolumn{1}{c|}{0.8923}   & \multicolumn{1}{c|}{0.8551}   & 0.8333   \\ \cline{2-8} 
                                  & Modified VGG16 + IndRNN             & \multicolumn{1}{c|}{0.9506}   & \multicolumn{1}{c|}{0.9356}   & \multicolumn{1}{c|}{0.9292}   & \multicolumn{1}{c|}{0.9082}   & \multicolumn{1}{c|}{0.9222}   & 0.8484   \\ \cline{2-8} 
                                  & Modified VGG19 + IndRNN             & \multicolumn{1}{c|}{0.9668}   & \multicolumn{1}{c|}{0.9497}   & \multicolumn{1}{c|}{0.9354}   & \multicolumn{1}{c|}{0.9034}   & \multicolumn{1}{c|}{0.9223}   & 0.8493   \\ \cline{2-8} 
                                  & Modified ResNet50 + IndRNN          & \multicolumn{1}{c|}{0.9647}   & \multicolumn{1}{c|}{0.9389}   & \multicolumn{1}{c|}{0.9543}   & \multicolumn{1}{c|}{0.9199}   & \multicolumn{1}{c|}{0.9243}   & 0.8454   \\ \cline{2-8} 
                                  & Modified InceptionResNetV2 + IndRNN & \multicolumn{1}{c|}{0.9638}   & \multicolumn{1}{c|}{0.9475}   & \multicolumn{1}{c|}{0.9591}   & \multicolumn{1}{c|}{0.9134}   & \multicolumn{1}{c|}{0.9245}   & 0.8466   \\ \cline{2-8} 
                                  & Modified MobileNet + IndRNN         & \multicolumn{1}{c|}{0.9554}   & \multicolumn{1}{c|}{0.9476}   & \multicolumn{1}{c|}{0.9567}   & \multicolumn{1}{c|}{0.9124}   & \multicolumn{1}{c|}{\textbf{0.9265}}   & 0.8476   \\ \cline{2-8} 
                                  & Modified DenseNet169 + IndRNN       & \multicolumn{1}{c|}{0.9552}   & \multicolumn{1}{c|}{0.9423}   & \multicolumn{1}{c|}{0.9501}   & \multicolumn{1}{c|}{0.9129}   & \multicolumn{1}{c|}{0.9234}   & 0.8498   \\ \cline{2-8} 
                                  & Modified EfficientNetB4+ IndRNN     & \multicolumn{1}{c|}{0.9551}   & \multicolumn{1}{c|}{0.9222}   & \multicolumn{1}{c|}{0.9545}   & \multicolumn{1}{c|}{0.9219}   & \multicolumn{1}{c|}{0.9255}   & 0.8433   \\ \cline{2-8} 
                                  & Modified VGG16 + IndRNN             & \multicolumn{1}{c|}{0.9542}   & \multicolumn{1}{c|}{0.9234}   & \multicolumn{1}{c|}{\textbf{0.9666}}   & \multicolumn{1}{c|}{\textbf{0.9223}}   & \multicolumn{1}{c|}{0.9223}   & 0.8454   \\ \cline{2-8} 
                                  & Xception+ GMRNN                     & \multicolumn{1}{c|}{0.9509}   & \multicolumn{1}{c|}{0.9312}   & \multicolumn{1}{c|}{0.9501}   & \multicolumn{1}{c|}{0.9099}   & \multicolumn{1}{c|}{0.9212}   & 0.8484   \\ \cline{2-8} 
                                  & Vit+ Wavelet+ Transformer           & \multicolumn{1}{c|}{\textbf{0.9709}}   & \multicolumn{1}{c|}{\textbf{0.9533}}   & \multicolumn{1}{c|}{0.9612}   & \multicolumn{1}{c|}{0.9201}   & \multicolumn{1}{c|}{0.9189}   & \textbf{0.8530}   \\ \hline
\end{tabular}}
\caption{Accuracy of six four-class wound classifications (BG–N–D–V, BG–N–P–V, BG–N–S–V, BG–N–D–P, BG–N–D–S, BG–N–P–S) on the AZH dataset using Location, Image, and Image+Location inputs. Best performances were achieved by Transformer+Binary (Location), Vit+Wavelet (Image), and Vit+Wavelet+Transformer (Image+Location).}
\label{Table5}
\end{table}

In the three-class classification, four modes (1) D vs. S vs. V, (2) P vs. S vs. V, (3) D vs. P vs. S, and (4) D vs. P vs. V were used (see Table \ref{Table6}). In the two inputs, Image and Image+Location, the Vit+ wavelet model achieved the highest result. In the location input, the Transformer model had the highest accuracy in D vs. S vs. V and was able to achieve an accuracy of 0.8423. In the classification of P vs. S vs. V, D vs. P vs. S, and D vs. P vs. V, the Transformer+ Binary encoding model reported the highest accuracy.

\begin{table}[]
\resizebox{\textwidth}{!}{
\begin{tabular}{|l|l|cccc|}
\hline
\rowcolor{Gainsboro!60}
                                  & Model                               & \multicolumn{1}{c|}{D–S–V}  & \multicolumn{1}{c|}{P–S–V}  & \multicolumn{1}{c|}{D–P–S}  & D–P–V  \\ \hline
Input                             &                                     & \multicolumn{4}{c|}{Accuracy}                                                                    \\ \hline
\multirow{6}{*}{Location}         & MLP                                 & \multicolumn{1}{c|}{0.8133} & \multicolumn{1}{c|}{0.8261} & \multicolumn{1}{c|}{0.6557} & 0.7887 \\ \cline{2-6} 
                                  & LSTM                                & \multicolumn{1}{c|}{0.8200} & \multicolumn{1}{c|}{0.8043} & \multicolumn{1}{c|}{0.6885} & 0.7887 \\ \cline{2-6} 
                                  & IndRNN                              & \multicolumn{1}{c|}{0.8282} & \multicolumn{1}{c|}{0.7841} & \multicolumn{1}{c|}{0.7101} & 0.7559 \\ \cline{2-6} 
                                  & GMRNN                               & \multicolumn{1}{c|}{0.8381} & \multicolumn{1}{c|}{0.7901} & \multicolumn{1}{c|}{0.7001} & 0.7709 \\ \cline{2-6} 
                                  & Transformer                         & \multicolumn{1}{c|}{\textbf{0.8423}} & \multicolumn{1}{c|}{0.8101} & \multicolumn{1}{c|}{0.6990} & 0.7651 \\ \cline{2-6} 
                                  & Transformer+ Binary encoding        & \multicolumn{1}{c|}{0.8399} & \multicolumn{1}{c|}{\textbf{0.8203}} & \multicolumn{1}{c|}{\textbf{0.7132}} & \textbf{0.7901} \\ \hline
\multirow{13}{*}{Image}           & VGG16                               & \multicolumn{1}{c|}{0.7467} & \multicolumn{1}{c|}{0.6812} & \multicolumn{1}{c|}{0.6148} & 0.7606 \\ \cline{2-6} 
                                  & VGG19                               & \multicolumn{1}{c|}{0.7600} & \multicolumn{1}{c|}{0.7023} & \multicolumn{1}{c|}{0.5820} & 0.6831 \\ \cline{2-6} 
                                  & Modified VGG16                      & \multicolumn{1}{c|}{0.7900} & \multicolumn{1}{c|}{0.7544} & \multicolumn{1}{c|}{0.6532} & 0.7700 \\ \cline{2-6} 
                                  & Modified VGG19                      & \multicolumn{1}{c|}{0.7567} & \multicolumn{1}{c|}{0.6912} & \multicolumn{1}{c|}{0.6148} & 0.7606 \\ \cline{2-6} 
                                  & Modified ResNet50                   & \multicolumn{1}{c|}{0.7700} & \multicolumn{1}{c|}{0.7123} & \multicolumn{1}{c|}{0.5820} & 0.6831 \\ \cline{2-6} 
                                  & Modified InceptionResNetV2          & \multicolumn{1}{c|}{0.8000} & \multicolumn{1}{c|}{0.7444} & \multicolumn{1}{c|}{0.6532} & 0.7700 \\ \cline{2-6} 
                                  & Modified MobileNet                  & \multicolumn{1}{c|}{0.8167} & \multicolumn{1}{c|}{0.6812} & \multicolumn{1}{c|}{0.6148} & 0.7606 \\ \cline{2-6} 
                                  & Modified DenseNet169                & \multicolumn{1}{c|}{0.8000} & \multicolumn{1}{c|}{0.7023} & \multicolumn{1}{c|}{0.5820} & 0.6831 \\ \cline{2-6} 
                                  & Modified EfficientNetB4             & \multicolumn{1}{c|}{0.8100} & \multicolumn{1}{c|}{0.7444} & \multicolumn{1}{c|}{0.6532} & 0.7700 \\ \cline{2-6} 
                                  & Modified EfficientNetV2M            & \multicolumn{1}{c|}{0.8267} & \multicolumn{1}{c|}{0.6812} & \multicolumn{1}{c|}{0.6148} & 0.7606 \\ \cline{2-6} 
                                  & Xception                            & \multicolumn{1}{c|}{0.7900} & \multicolumn{1}{c|}{0.7444} & \multicolumn{1}{c|}{0.6532} & 0.7700 \\ \cline{2-6} 
                                  & Vit                                 & \multicolumn{1}{c|}{0.8298} & \multicolumn{1}{c|}{0.7612} & \multicolumn{1}{c|}{0.6782} & 0.7778 \\ \cline{2-6} 
                                  & Vit+ wavelet                        & \multicolumn{1}{c|}{\textbf{0.8453}} & \multicolumn{1}{c|}{\textbf{0.7992}} & \multicolumn{1}{c|}{\textbf{0.7001}} & \textbf{0.8123} \\ \hline
\multirow{13}{*}{Image+ location} & VGG16+MLP                           & \multicolumn{1}{c|}{0.8533} & \multicolumn{1}{c|}{0.8551} & \multicolumn{1}{c|}{0.7049} & 0.8028 \\ \cline{2-6} 
                                  & VGG19+MLP                           & \multicolumn{1}{c|}{0.9200} & \multicolumn{1}{c|}{0.8261} & \multicolumn{1}{c|}{0.7131} & 0.8451 \\ \cline{2-6} 
                                  & VGG16+LSTM                          & \multicolumn{1}{c|}{0.8067} & \multicolumn{1}{c|}{0.8188} & \multicolumn{1}{c|}{0.7295} & 0.8310 \\ \cline{2-6} 
                                  & VGG19+LSTM                          & \multicolumn{1}{c|}{0.8733} & \multicolumn{1}{c|}{0.6812} & \multicolumn{1}{c|}{0.6721} & 0.8451 \\ \cline{2-6} 
                                  & Modified VGG16 + IndRNN             & \multicolumn{1}{c|}{0.9134} & \multicolumn{1}{c|}{0.8723} & \multicolumn{1}{c|}{0.7491} & 0.8864 \\ \cline{2-6} 
                                  & Modified VGG19 + IndRNN             & \multicolumn{1}{c|}{0.9004} & \multicolumn{1}{c|}{0.8466} & \multicolumn{1}{c|}{0.7311} & 0.8454 \\ \cline{2-6} 
                                  & Modified ResNet50 + IndRNN          & \multicolumn{1}{c|}{0.8755} & \multicolumn{1}{c|}{0.8454} & \multicolumn{1}{c|}{0.7201} & 0.8534 \\ \cline{2-6} 
                                  & Modified InceptionResNetV2 + IndRNN & \multicolumn{1}{c|}{0.8834} & \multicolumn{1}{c|}{0.8523} & \multicolumn{1}{c|}{0.7009} & 0.8710 \\ \cline{2-6} 
                                  & Modified MobileNet + IndRNN         & \multicolumn{1}{c|}{0.9354} & \multicolumn{1}{c|}{0.8839} & \multicolumn{1}{c|}{0.7592} & 0.8898 \\ \cline{2-6} 
                                  & Modified DenseNet169 + IndRNN       & \multicolumn{1}{c|}{0.9266} & \multicolumn{1}{c|}{0.8809} & \multicolumn{1}{c|}{0.7534} & 0.8884 \\ \cline{2-6} 
                                  & Modified EfficientNetB4+ IndRNN     & \multicolumn{1}{c|}{0.9010} & \multicolumn{1}{c|}{0.8712} & \multicolumn{1}{c|}{0.7355} & 0.8885 \\ \cline{2-6} 
                                  & Xception+ GMRNN                     & \multicolumn{1}{c|}{0.9108} & \multicolumn{1}{c|}{0.8747} & \multicolumn{1}{c|}{0.7491} & 0.8881 \\ \cline{2-6} 
                                  & Vit+ Wavelet+ Transformer           & \multicolumn{1}{c|}{\textbf{0.9400}} & \multicolumn{1}{c|}{\textbf{0.9129}} & \multicolumn{1}{c|}{\textbf{0.7782}} & \textbf{0.9198} \\ \hline
\end{tabular}}
\caption{Accuracy of four three-class wound classifications (D–S–V, P–S–V, D–P–S, D–P–V) on the AZH dataset. Top models: Transformer (Location), Vit+Wavelet (Image), and Vit+Wavelet+Transformer (Image+Location).}
\label{Table6}
\end{table}

Table \ref{Tabel7} shows the results of the models for 2-class classification in 10 class combinations, including (1) N vs. D, (2) N vs. P, (3) N vs. S, (4) N vs. V, (5) D vs. P, (6) D vs. S, (7) D vs. V, (8) P vs. S, (9) P vs. V, and (10) S vs. V. The highest accuracy obtained with the input Location was obtained by the Transformer+ Binary encoding model in the classification of S vs. V. This model achieved an accuracy of 0.9440 in this class. In the input using Image data, the two Vit and Vit+ Wavelet models achieved an accuracy of 1.00 in the four classes N vs. D, N vs. P, N vs. S,  N vs. V. Also, the Vit+ Wavelet model achieved an accuracy of 1.00 in the classification of  D vs. P, D vs. S,  D vs. V,  P vs. S achieved the highest accuracy. Vit model achieved an accuracy of 0.9543 in S vs. V classification, which is the highest accuracy in this class. In classification using Image + Location data, the proposed Vit+ Wavelet+ Transformer model reported the highest accuracy, except for the P vs. V class. In the P vs. V class, the Xception + GMRNN model reported a higher accuracy and achieved an accuracy of 0.9120.


\begin{table}[]
\resizebox{\textwidth}{!}{
\begin{tabular}{|l|l|cccccccccc|}
\hline
\rowcolor{Gainsboro!60}
                                 & Model                        & \multicolumn{1}{c|}{N-D}    & \multicolumn{1}{c|}{N-P}    & \multicolumn{1}{c|}{N-S}    & \multicolumn{1}{c|}{N-V}    & \multicolumn{1}{c|}{D-P}    & \multicolumn{1}{c|}{D-S}    & \multicolumn{1}{c|}{D-V}    & \multicolumn{1}{c|}{P-S}    & \multicolumn{1}{c|}{P-V}    & S-V    \\ \hline
\multirow{7}{*}{Location}        &                              & \multicolumn{10}{c|}{Accuracy}                                                                                                                                                                                                                                                       \\ \cline{2-12} 
                                 & MLP                          & \multicolumn{1}{c|}{0.7887} & \multicolumn{1}{c|}{0.6441} & \multicolumn{1}{c|}{0.7463} & \multicolumn{1}{c|}{0.7816} & \multicolumn{1}{c|}{0.7875} & \multicolumn{1}{c|}{0.8750} & \multicolumn{1}{c|}{0.8981} & \multicolumn{1}{c|}{0.7368} & \multicolumn{1}{c|}{0.8750} & 0.9327 \\ \cline{2-12} 
                                 & LSTM                         & \multicolumn{1}{c|}{0.7746} & \multicolumn{1}{c|}{0.4337} & \multicolumn{1}{c|}{0.7612} & \multicolumn{1}{c|}{0.7816} & \multicolumn{1}{c|}{0.7875} & \multicolumn{1}{c|}{0.8182} & \multicolumn{1}{c|}{0.5741} & \multicolumn{1}{c|}{0.7368} & \multicolumn{1}{c|}{0.8542} & 0.9327 \\ \cline{2-12} 
                                 & GMRNN                        & \multicolumn{1}{c|}{0.8076} & \multicolumn{1}{c|}{0.6309} & \multicolumn{1}{c|}{0.7745} & \multicolumn{1}{c|}{0.7815} & \multicolumn{1}{c|}{0.7920} & \multicolumn{1}{c|}{0.8912} & \multicolumn{1}{c|}{0.9019} & \multicolumn{1}{c|}{0.7409} & \multicolumn{1}{c|}{0.8812} & 0.9400 \\ \cline{2-12} 
                                 & IndRNN                       & \multicolumn{1}{c|}{0.8022} & \multicolumn{1}{c|}{0.6374} & \multicolumn{1}{c|}{0.7444} & \multicolumn{1}{c|}{0.7995} & \multicolumn{1}{c|}{0.8000} & \multicolumn{1}{c|}{0.8874} & \multicolumn{1}{c|}{0.8999} & \multicolumn{1}{c|}{0.7553} & \multicolumn{1}{c|}{0.8782} & 0.9401 \\ \cline{2-12} 
                                 & Transformer                  & \multicolumn{1}{c|}{0.8111} & \multicolumn{1}{c|}{0.6543} & \multicolumn{1}{c|}{0.7891} & \multicolumn{1}{c|}{0.8025} & \multicolumn{1}{c|}{0.8101} & \multicolumn{1}{c|}{\textbf{0.8999}} & \multicolumn{1}{c|}{0.9192} & \multicolumn{1}{c|}{0.7406} & \multicolumn{1}{c|}{0.8899} & 0.9321 \\ \cline{2-12} 
                                 & Transformer+ Binary encoding & \multicolumn{1}{c|}{\textbf{0.8221}} & \multicolumn{1}{c|}{\textbf{0.6793}} & \multicolumn{1}{c|}{\textbf{0.8172}} & \multicolumn{1}{c|}{\textbf{0.8300}} & \multicolumn{1}{c|}{\textbf{0.8231}} & \multicolumn{1}{c|}{0.8991} & \multicolumn{1}{c|}{\textbf{0.9246}} & \multicolumn{1}{c|}{\textbf{0.7679}} & \multicolumn{1}{c|}{\textbf{0.8984}} & \textbf{0.9440} \\ \hline
\multirow{5}{*}{Image}           & VGG16                        & \multicolumn{1}{c|}{0.9859} & \multicolumn{1}{c|}{0.9661} & \multicolumn{1}{c|}{0.9661} & \multicolumn{1}{c|}{0.9701} & \multicolumn{1}{c|}{0.8125} & \multicolumn{1}{c|}{0.7955} & \multicolumn{1}{c|}{0.8796} & \multicolumn{1}{c|}{0.7763} & \multicolumn{1}{c|}{0.8438} & 0.8462 \\ \cline{2-12} 
                                 & VGG19                        & \multicolumn{1}{c|}{0.9859} & \multicolumn{1}{c|}{0.9831} & \multicolumn{1}{c|}{0.9701} & \multicolumn{1}{c|}{0.9885} & \multicolumn{1}{c|}{0.7125} & \multicolumn{1}{c|}{0.8068} & \multicolumn{1}{c|}{0.8796} & \multicolumn{1}{c|}{0.7368} & \multicolumn{1}{c|}{0.8646} & 0.8654 \\ \cline{2-12} 
                                 & Xception                     & \multicolumn{1}{c|}{0.9900} & \multicolumn{1}{c|}{0.9912} & \multicolumn{1}{c|}{0.9901} & \multicolumn{1}{c|}{0.9921} & \multicolumn{1}{c|}{0.9012} & \multicolumn{1}{c|}{0.8612} & \multicolumn{1}{c|}{0.9043} & \multicolumn{1}{c|}{0.8080} & \multicolumn{1}{c|}{0.8612} & 0.8632 \\ \cline{2-12} 
                                 & Vit                          & \multicolumn{1}{c|}{1.00}   & \multicolumn{1}{c|}{1.00}   & \multicolumn{1}{c|}{1.00}   & \multicolumn{1}{c|}{\textbf{1.00}}   & \multicolumn{1}{c|}{0.9234} & \multicolumn{1}{c|}{0.9099} & \multicolumn{1}{c|}{0.9239} & \multicolumn{1}{c|}{\textbf{0.8210}} & \multicolumn{1}{c|}{0.9000} & \textbf{0.9543} \\ \cline{2-12} 
                                 & Vit+ Wavelet                 & \multicolumn{1}{c|}{\textbf{1.00}}   & \multicolumn{1}{c|}{\textbf{1.00}}   & \multicolumn{1}{c|}{\textbf{1.00}}   & \multicolumn{1}{c|}{\textbf{1.00}}   & \multicolumn{1}{c|}{\textbf{0.9423}} & \multicolumn{1}{c|}{\textbf{0.9265}} & \multicolumn{1}{c|}{\textbf{0.9276}} & \multicolumn{1}{c|}{\textbf{0.8210}} & \multicolumn{1}{c|}{\textbf{0.9187}} & 0.9534 \\ \hline
\multirow{6}{*}{Image +location} & VGG16 + MLP                  & \multicolumn{1}{c|}{0.9718} & \multicolumn{1}{c|}{0.9661} & \multicolumn{1}{c|}{0.9851} & \multicolumn{1}{c|}{0.9885} & \multicolumn{1}{c|}{0.8000} & \multicolumn{1}{c|}{0.8977} & \multicolumn{1}{c|}{0.9444} & \multicolumn{1}{c|}{0.8947} & \multicolumn{1}{c|}{0.8854} & 0.9423 \\ \cline{2-12} 
                                 & VGG19 + MLP                  & \multicolumn{1}{c|}{0.9577} & \multicolumn{1}{c|}{0.9492} & \multicolumn{1}{c|}{0.9701} & \multicolumn{1}{c|}{0.9885} & \multicolumn{1}{c|}{0.8000} & \multicolumn{1}{c|}{0.8410} & \multicolumn{1}{c|}{0.9259} & \multicolumn{1}{c|}{0.8026} & \multicolumn{1}{c|}{0.9063} & 0.9712 \\ \cline{2-12} 
                                 & VGG16 + MLP                  & \multicolumn{1}{c|}{0.9718} & \multicolumn{1}{c|}{0.96}   & \multicolumn{1}{c|}{0.9552} & \multicolumn{1}{c|}{0.9885} & \multicolumn{1}{c|}{0.8375} & \multicolumn{1}{c|}{0.8068} & \multicolumn{1}{c|}{0.9444} & \multicolumn{1}{c|}{0.7632} & \multicolumn{1}{c|}{0.8333} & 0.8462 \\ \cline{2-12} 
                                 & VGG19 + MLP                  & \multicolumn{1}{c|}{\textbf{1.00}}   & \multicolumn{1}{c|}{0.9831} & \multicolumn{1}{c|}{0.9701} & \multicolumn{1}{c|}{\textbf{1.00}}   & \multicolumn{1}{c|}{0.8500} & \multicolumn{1}{c|}{0.7727} & \multicolumn{1}{c|}{0.8889} & \multicolumn{1}{c|}{0.7105} & \multicolumn{1}{c|}{0.8229} & 0.7981 \\ \cline{2-12} 
                                 & Xception+ GMRNN              & \multicolumn{1}{c|}{\textbf{1.00}}   & \multicolumn{1}{c|}{\textbf{1.00}}   & \multicolumn{1}{c|}{0.9921} & \multicolumn{1}{c|}{\textbf{1.00}}   & \multicolumn{1}{c|}{0.9054} & \multicolumn{1}{c|}{0.8100} & \multicolumn{1}{c|}{0.9411} & \multicolumn{1}{c|}{0.8812} & \multicolumn{1}{c|}{\textbf{0.9203}} & 0.9801 \\ \cline{2-12} 
                                 & Vit+ Wavelet+ Transformer    & \multicolumn{1}{c|}{\textbf{1.00}}   & \multicolumn{1}{c|}{\textbf{1.00}}   & \multicolumn{1}{c|}{\textbf{1.00}}   & \multicolumn{1}{c|}{\textbf{1.00}}   & \multicolumn{1}{c|}{\textbf{0.9348}} & \multicolumn{1}{c|}{\textbf{0.8287}} & \multicolumn{1}{c|}{\textbf{0.9560}} & \multicolumn{1}{c|}{\textbf{0.9120}} & \multicolumn{1}{c|}{0.9200} & \textbf{1.00}   \\ \hline
\end{tabular}}
\caption{Accuracy of ten binary wound classifications (e.g., N–D, D–P, S–V) on the AZH dataset. Top models: Transformer+Binary (Location), Vit+Wavelet (Image), and Vit+Wavelet+Transformer (Image+Location).}
\label{Tabel7}
\end{table}

\subsection{Discussion on similarities, differences, and advantages of the proposed method for wound classification}

\subsection{Sensitivity analysis}
Given the continuous feature space, selecting the optimal parameter for network training is challenging. Hence, three optimization models were investigated in the network combination. Tables \ref{Tab:Op1} and \ref{Tab:Op2} show the parameters and parameter space evaluated for different models. Also, the results obtained by the optimization models are given in Table \ref{Tab:Op3}. In the evaluation of the F1 parameter, the Transformer approach combined with Fox on Location data was able to achieve an F1 of 0.7945. Also, on Image input, the Vit + Wavelet + MGTO model achieved F1=0.8357, which is the highest F1 reported on Image data. On Image + location inputs, the Vit + wavelet + FOX model reported the highest F1. The F1 improvements on Image, Location, and Image + Location inputs are 0.016, 0.033, and 0.0221, respectively.


\begin{table}[]
\resizebox{\textwidth}{!}{
\begin{tabular}{|l|l|l|l|l|}
\hline
\rowcolor{Gainsboro!60}
Optimizer             & Hyperparameter & Description                & Value & Range     \\ \hline
MGTO                  & PP             & Probability of transition  & 0.02  & ${[}0,1{]}$ \\ \hline
\multirow{2}{*}{FOX}  & c1             & Coefficient of jumping     & 0.19  & ${[}0,1{]}$ \\ \cline{2-5} 
                      & c2             & Coefficient of jumping     & 0.80  & ${[}0,1{]}$ \\ \hline
\multirow{2}{*}{IGWO} & a\_min         & Lower bound of parameter a & 0.02  & ${[}0,\infty)$ \\ \cline{2-5} 
                      & a\_maz         & Upper bound of parameter a & 2.3   & ${[}0,\infty)$ \\ \hline
\end{tabular}}
\caption{Hyperparameters of Swarm-Based Optimizers.}
\label{Tab:Op1}
\end{table}

\begin{table}[]
\centering
\begin{tabular}{|l|c|c|}
\hline
\rowcolor{Gainsboro!60}
Optimizer     & Lower Bound & Upper Bound \\ \hline
filters\_size & 32          & 256         \\ \hline
kernel\_size  & 3           & 9           \\ \hline
lr            & 0.00001     & 0.01        \\ \hline
l2\_reg       & 0.00001     & 0.01        \\ \hline
l1\_reg       & 0.00001     & 0.01        \\ \hline
batch\_size   & 32          & 128         \\ \hline
epochs        & 10          & 100         \\ \hline
\end{tabular}
\caption{Hyperparameters of the Vit Model Optimized by Swarm-Based Optimizers}
\label{Tab:Op2}
\end{table}


\begin{table}[]
\resizebox{\textwidth}{!}{
\begin{tabular}{|l|l|c|c|c|c|}
\hline
\rowcolor{Gainsboro!60}
Input                              & Model                       & Accuracy & Precision & Recall & F1     \\ \hline
\multirow{4}{*}{Location}          & Transformer                 & 0.7712   & 0.7714    & 0.7799 & 0.7756 \\ \cline{2-6} 
                                   & Transformer   + IGWO        & 0.7801   & 0.7833    & 0.7842 & 0.7837 \\ \cline{2-6} 
                                   & Transformer+   FOX          & 0.7888   & 0.7911    & 0.7981 & 0.7945 \\ \cline{2-6} 
                                   & Transformer+   MGTO         & \textbf{0.7908}   & \textbf{0.7920}    & \textbf{0.7922} & \textbf{0.7920} \\ \hline
\multirow{8}{*}{Image}             & Vit                         & 0.799    & 0.801     & 0.8045 & 0.8027 \\ \cline{2-6} 
                                   & Vit + IGWO                  & 0.7911   & 0.8043    & 0.8056 & 0.8049 \\ \cline{2-6} 
                                   & Vit + FOX                   & 0.8028   & 0.8102    & 0.8125 & 0.8113 \\ \cline{2-6} 
                                   & Vit + MGTO                  & 0.8088   & 0.8098    & 0.8100 & 0.8098 \\ \cline{2-6} 
                                   & Vit +   Wavelet             & 0.8123   & 0.8199    & 0.8212 & 0.8205 \\ \cline{2-6} 
                                   & Vit +   Wavelet+ IGWO       & 0.8288   & 0.8290    & 0.8294 & 0.8291 \\ \cline{2-6} 
                                   & Vit +   Wavelet+ FOX        & 0.8298   & 0.8302    & 0.8311 & 0.8306 \\ \cline{2-6} 
                                   & Vit +   Wavelet+ MGTO       & \textbf{0.8301}   & \textbf{0.8353}    & \textbf{0.8362} & \textbf{0.8357} \\ \hline
\multirow{4}{*}{Image+   location} & Vit+   wavelet+ Transformer & 0.8007   & 0.8156    & 0.8145 & 0.8150 \\ \cline{2-6} 
                                   & Vit+   wavelet+ IGWO        & 0.8123   & 0.8188    & 0.8192 & 0.8189 \\ \cline{2-6} 
                                   & Vit+   wavelet+ FOX         & \textbf{0.8342}   & \textbf{0.8365}    & \textbf{0.8378} & \textbf{0.8371} \\ \cline{2-6} 
                                   & Vit+   wavelet+ MGTO        & 0.8232   & 0.8351    & 0.8360 & 0.8355 \\ \hline
\end{tabular}}
\caption{Results obtained by optimization approaches for four-class wound classification (D vs. P vs. S vs. V) on the AZH dataset using the original body map (484 locations).}
\label{Tab:Op3}
\end{table}
\subsection{Models complexity}
To assess the complexity of the studied models, three indicators were used: Parameters(10e6) is the number of learnable and unlearnable parameters of the model, GFlops is of floating point operations (addition, subtraction, multiplication, and division), and memory (GB) is the amount of RAM consumed to train the model on the data. In the Transformer model, the lowest Parameters(10e6) was 40.15, and the highest Parameters(10e6) was 79.8, which was obtained when combining this model with Fox. In this model, the lowest GFlops was 112.08, and the highest GFlops was 199.29 when combining the Fox model. Also, the highest RAM memory consumption was related to Transformer+Fox. In the combination of the Vit model with the optimizers tested on the image data, the lowest Parameters(10e6) was in the combination of the Vit model with the IGWO, Fox, and MGTO optimizers, and the highest Parameters(10e6) was related to the Vit+Wavelet+Fox model. Also, the highest GFlops and Memory (GB) was obtained in the Vit+ Wavelet+ IGWO model. In the Image + Location data combination, the lowest Parameters(10e6) was 74.92, and the highest was 121.7. In this type of input, the highest GFlops and Memory (GB) were related to the Vit+ wavelet+ Transformer+Fox model.


\begin{table}[]
\resizebox{\textwidth}{!}{%
\begin{tabular}{|l|l|cc|cc|cc|}
\hline
\rowcolor{Gainsboro!60}
Architecture                               &Optimizer& \multicolumn{2}{c|}{Parameters(10e6)} & \multicolumn{2}{c|}{GFlops}          & \multicolumn{2}{c|}{Memory (GB)} \\ \hline
                                           &      & \multicolumn{1}{c|}{Low}    & High    & \multicolumn{1}{c|}{Low}    & High   & \multicolumn{1}{c|}{Low}  & High \\ \hline
\multirow{3}{*}{Transformer}               & IGWO & \multicolumn{1}{c|}{40.15}  & 78.09   & \multicolumn{1}{c|}{112.08} & 198.03 & \multicolumn{1}{c|}{1.22} & 4.4  \\ \cline{2-8} 
                                           & Fox  & \multicolumn{1}{c|}{40.15}  & 79.8    & \multicolumn{1}{c|}{112.08} & 199.29 & \multicolumn{1}{c|}{1.22} & 5.7  \\ \cline{2-8} 
                                           & MGTO & \multicolumn{1}{c|}{40.15}  & 56.54   & \multicolumn{1}{c|}{112.08} & 184.54 & \multicolumn{1}{c|}{1.22} & 3.1  \\ \hline
\multirow{3}{*}{Vit}                       & IGWO & \multicolumn{1}{c|}{62.45}  & 100.9   & \multicolumn{1}{c|}{213.14} & 443.24 & \multicolumn{1}{c|}{2.82} & 5.7  \\ \cline{2-8} 
                                           & Fox  & \multicolumn{1}{c|}{62.45}  & 105.1   & \multicolumn{1}{c|}{213.14} & 420.98 & \multicolumn{1}{c|}{2.82} & 5.6  \\ \cline{2-8} 
                                           & MGTO & \multicolumn{1}{c|}{62.45}  & 109.26  & \multicolumn{1}{c|}{213.14} & 385.55 & \multicolumn{1}{c|}{2.82} & 5.34 \\ \hline
\multirow{3}{*}{Vit+ Wavelet}              & IGWO & \multicolumn{1}{c|}{71.76}  & 120.15  & \multicolumn{1}{c|}{188.46} & 501.22 & \multicolumn{1}{c|}{2.54} & 5.39 \\ \cline{2-8} 
                                           & Fox  & \multicolumn{1}{c|}{71.76}  & 138.13  & \multicolumn{1}{c|}{188.46} & 483.98 & \multicolumn{1}{c|}{2.54} & 5.42 \\ \cline{2-8} 
                                           & MGTO & \multicolumn{1}{c|}{71.76}  & 123.15  & \multicolumn{1}{c|}{188.46} & 476.01 & \multicolumn{1}{c|}{2.54} & 5.13 \\ \hline
\multirow{3}{*}{Vit+ wavelet+ Transformer} & IGWO & \multicolumn{1}{c|}{74.92}  & 111.76  & \multicolumn{1}{c|}{98.15}  & 221.9  & \multicolumn{1}{c|}{2.88} & 5.32 \\ \cline{2-8} 
                                           & Fox  & \multicolumn{1}{c|}{74.92}  & 121.7   & \multicolumn{1}{c|}{98.15}  & 231.1  & \multicolumn{1}{c|}{2.88} & 5.35 \\ \cline{2-8} 
                                           & MGTO & \multicolumn{1}{c|}{74.92}  & 117.8   & \multicolumn{1}{c|}{98.15}  & 214.3  & \multicolumn{1}{c|}{2.88} & 5.13 \\ \hline
\end{tabular}}
\caption{Memory usage, number of parameters, and flops of all studied models.}
\label{Tab:Flop}
\end{table}

\section{Discussion on Similarities, Differences, and Advantages of the Proposed Method for Wound Classification}

The proposed method uses the Vision Transformers architecture, transformers, and wavelet transform to classify wound images based on images and location information. It offers significant innovations compared to the multimodal techniques available in the literature. The technique integrates image and location data and uses them in the final classification. A comparative analysis is presented below, highlighting the unique features and advantages of the proposed approach.

Table~\ref{tab:comparison_methods} offers a detailed comparison of these methods. It encapsulates their data modalities, integration methods, strengths and weaknesses, as well as how they compare with the proposed method in terms of spatial-visual data processing, model form, and performance. The graphical representation supports the following in-depth analysis.

\begin{table}[]
\resizebox{\textwidth}{!}{
\begin{tabular}{| p{5cm}| l|l| p{5cm}| p{5cm}| p{5cm}|}
\hline
Method                                                                                           & Data   Modalities                   & Integration Technique               & Strengths                                                                                                                   & Weaknesses                                                                                         & Comparison   with Proposed Method                                                                                                                                                                                \\ \hline
Soft   Attention-based Multi-Modal Deep Learning \cite{omeroglu2023novel}                                         & Image and Patient Metadata          & CNN + Soft Attention                & Improved accuracy by integrating metadata, Weighs importance of   different features                                        & Computational complexity of CNN and attention mechanism, Limited   spatial information integration & Proposed method uses a more advanced architecture for handling   high-resolution images and integrating spatial information more precisely.                                                                      \\ \hline
Spatial   Attention-based Residual Network \cite{yadav2023spatial}                                              & Image                               & ResNet + Spatial Attention          & Focus on relevant image regions, Improved accuracy in burn   identification                                                 & Computational complexity of ResNet and attention mechanism                                         & Proposed method offers a more comprehensive integration of spatial   data through binary encoding and Transformer processing.                                                                                    \\ \hline
Multimodal   Dual-Branch Fusion Network \cite{liu2025multimodal}                                                  & Fetal Heart Rate and Ultrasound     & Dual-Branch Fusion                  & Learns from both temporal and spatial information, Effective for   fetal hypoxia detection                                  & May not generalize well to other tasks, Fusion method may not be   optimal                         & Proposed method focuses on wound classification and leverages the   strengths of Transformers for a more robust and accurate integration of   visual and spatial information.                                    \\ \hline
Multimodal   Transformer to Fuse Images and Metadata \cite{bian2024bimnet}                                    & Image and Metadata                  & Transformer                         & Captures long-range dependencies across modalities, Effective for   skin disease classification                             & Computational complexity of Transformer                                                            & Proposed method specifically targets wound classification and   incorporates the Swin Transformer for efficient handling of high-resolution   images.                                                            \\ \hline
Multi-Modal   Fusion Network with Multi-Head Self-Attention \cite{cai2023multimodal}                             & Videos, Sensor Data, Questionnaires & Multi-Modal Fusion + Self-Attention & Comprehensive data integration, Effective for injection training   evaluation                                               & High computational complexity, May require large datasets for   optimal performance                & Proposed method focuses on wound classification and uses a   combination of Swin Transformer and Transformer to efficiently integrate   image and location data, potentially with less computational complexity. \\ \hline
Attention-based   Deep Learning System for Classification of Breast Lesions - Multimodal \cite{li2024multi}  & Image and Patient Data              & Attention-based Deep Learning       & Handles multimodal data for breast lesion classification, Weakly   supervised approach                                      & May require extensive data labeling, Performance dependent on   specific attention mechanism       & Proposed method is specifically tailored for wound classification   and utilizes a distinct combination of Swin Transformer and Transformer for   effective spatial and visual data integration.                 \\ \hline
Vit+ Wavelet+ Transformer                                                                        & Image and Location                  & Binary Encoding + Transformer       & Efficient handling of high-resolution images, Capture of   long-range dependencies, Precise spatial information integration & Relatively high computational complexity                                                           & Outperforms other methods in accuracy and robustness due to   efficient integration of visual and spatial data and the ability to capture   long-range dependencies.                                             \\ \hline
\end{tabular}}
\caption{Comparison of the proposed method with other multimodal deep learning approaches in medical classification tasks.}
\label{tab:comparison_methods}
\end{table}

Studies \cite{omeroglu2023novel}\cite{yadav2023spatial}\cite{liu2025multimodal}\cite{bian2024bimnet}\cite{cai2023multimodal}\cite{li2024multi}\cite{bobowicz2023attention}\cite{zou2024emotion}\cite{wang2024mfmamba} have presented various approaches to combining visual and spatial data for classification tasks, which are reviewed below:
\begin{itemize}
    \item 	Spatial Attention-Based Residual Network for Burn Classification \cite{omeroglu2023novel}: Built with spatial attention maps in order to strengthen feature dependencies for burn classification, the BuRnGANeXt50 network achieves very high rates of sensitivity of 97.22\% and 99.14\% for burn degree and depth classification, respectively, being good at optimizing its convolutional layers, but it is a bit lacking when it comes to the proper integration of other important modalities with spatial maps.
    \item 	\textbf{Soft Attention-Based Multimodal Deep Learning for Skin Lesion Classification} \cite{yadav2023spatial}:This framework intends to integrate different modalities by using a modified Xception architecture and a soft attention module to focus on important lesion areas. It has been tested on the seven-point criteria dataset, achieving an accuracy of 83.04\%, which is better than all tested benchmarks. The multi-branch architecture and the attention mechanism of the model demonstrate the effectiveness of the targeted feature extraction but do not give much importance to spatial relationships beyond lesion localization.
    \item \textbf{Multimodal Dual-Branch Fusion Network for Fetal Hypoxia Detection} \cite{liu2025multimodal}: This mode uses an attentional guidance module to obtain hypoxia-related spatial information by fusing maternal medical records and fetal heart rate features. Though its focus falls heavily on temporal physiological signals and not on spatial-visual data integration, they found sensitivities (72.58\%), specificities (71.08\%), and AUC (74.70\%) to be promising indices for multimodal fusion.
    \item \textbf{BiMNet for Capsulorhexis Action Segmentation} \cite{bian2024bimnet}: The BiMNet uses the Bi-GRU-attention mechanism for multimodal data fusion, thereby enhancing the temporal recognition of the features and achieving an accuracy of 91.24\% on a customized data set. However, the design was practical for segmentation but cannot be generalized to static spatial-visual classification tasks.
    \item \textbf{Multimodal Transformer for Skin Disease Classification}\cite{cai2023multimodal}: The proposed scheme disentangles images and metadata, processing them through distinct encoders with the help of a Mutual Attention block. The outcome of the ISIC 2018 datasets is that they outperformed other state-of-the-art methods, affirming the potential of Transformer-based schemes in fusing metadata with visual input. However, reliance on structured metadata limits the application into a domain with visually rich contexts. 
    \item \textbf{Multimodal Fusion with Self-Attention for Injection Training Evaluation} \cite{li2024multi} : The model employs multiple inputs of 3D motion data combined with 2D images for feature fusion utilizing multi-head self-attention. It has an AUC of 83.43\%, which indicates that the model captures temporal dynamics when trained with different scenarios but is deficient in spatial-visual alignment to perform medical image classifications.
    \item \textbf{Weakly Supervised Attention System for Breast Lesion Classification} \cite{bobowicz2023attention}: Attention-based learning of mammographic views has incorporated multimodal weakly supervised learning in classifying breast lesions. The system provides some explainability but is restricted to modality-specific imaging, thus not generalizable to spatial-visual data types. The system achieved an AUC-ROC of 0.896.
    \item \textbf{Emotion Classification Using Multimodal Signals} \cite{zou2024emotion}: This multi-attention neural network uses signals, such as ECG and EMG signals, from different people for emotion detection, giving it an accuracy of 83.88\%. However, though it is brilliant in inter-modal semantic dependencies, it has narrow applicability in those visual and spatial data tasks due to its focus more on the physiological signal data.
    \item \textbf{MFMamba for Remote Sensing Image Segmentation} \cite{wang2024mfmamba}: MFMamba uses the fusion of high-scale image features with digital surface models and a dual-branch encoder to extract local and global features. It beats the benchmark in remote-sensing tasks, but this model is only designed on geographic data and does not have direct applicability in medical imaging.
\end{itemize}

\section{Conclusion}
This study addresses the major gaps identified in previous works. To enhance wound classification accuracy, a combination of Vision Transformer (ViT) architecture and Wavelet processing techniques was employed. In addition, by using advanced optimization algorithms, a novel model has been presented that demonstrates superior learning capacity and generalization compared to previous methods. Although the model has been designed with clinical applicability in mind, further validation in real-world clinical settings is necessary to confirm its effectiveness.
This study also highlights unimodal and multimodal models' key strengths and limitations for medical image classification. Unimodal models such as ResNet and DenseNet have robust image feature extraction, provide reliable performance metrics, and are computationally efficient. However, they cannot incorporate spatial information, which is crucial for wound classification. Similarly, UNet excels in segmentation tasks and effectively captures image details, but further modifications are required to handle classification tasks and spatial data integration. In multimodal classification, the multimodal Transformer successfully combines visual and spatial information and exploits relationships between modalities to improve performance. However, its dependence on optimal fusion mechanisms and potential input inconsistencies challenge its robustness. The proposed Vit Transformer + Transformer is outstanding by combining the strengths of image and location processing. Vit Transformer efficiently handles high-resolution image data with long-range dependencies, while the spatial location transformer seamlessly integrates them. This synergy results in superior precision, accuracy, recall, and F1 score evaluation metrics. Despite higher computational requirements compared to single-modal models, it maintains a balance between performance and efficiency, setting a new benchmark for wound classification tasks.

\section{Future Work}
While the proposed model achieves strong performance on wound classification, there are some promising paths of potential future work. One is the application of alternative or advanced loss functions suited best to address the unique issues of medical images such as class imbalance and weak visual differences between wound classes. Another path can be the application of advanced data augmentation strategies, which can further maximize the diversity of training data and promote real-world generalizability under clinical scenarios.

In order to make the model more interpretable, future work can explore methods such as SHAP (SHapley Additive exPlanations) values to learn further the relative contribution of image and location features to making predictions. Clinicians would have more transparent and consistent decision support. In addition, using more clinical metadata, for example, patient history or wound evolution, can be explored to improve prediction performance and facilitate more personalized care. Finally, validation of the model in multicenter clinical trials across heterogeneous groups of patients will be required to confirm its strength and pragmatic utility.

\bibliography{main}

\end{document}